\documentclass[conference]{IEEEtran}
\usepackage{times}

\usepackage[numbers]{natbib}
\usepackage{multicol}
\usepackage[bookmarks=true]{hyperref}
\usepackage[dvipsnames]{xcolor}
\usepackage{graphicx}
\usepackage{subcaption}
\usepackage{booktabs}
\usepackage{adjustbox}
\usepackage{makecell}
\usepackage{multirow}
\usepackage{diagbox}
\usepackage{amsmath,amssymb}
\hypersetup{
    colorlinks=true,
    linkcolor=orange,
    filecolor=magenta,
    urlcolor=orange,
    citecolor=orange,
}
\usepackage[capitalise, nameinlink]{cleveref}
\usepackage{algorithm}
\usepackage{algpseudocode}
\usepackage{stfloats}
\usepackage{wrapfig}

\begin{document}

\title{Efficient Data Collection for Robotic Manipulation via Compositional Generalization}

\author{Jensen Gao$^1$,
Annie Xie$^1$,
Ted Xiao$^2$,
Chelsea Finn$^{1,2}$,
Dorsa Sadigh$^{1,2}$ \\
$^1$Stanford University, $^2$Google DeepMind}

\maketitle

\begin{abstract}
Data collection has become an increasingly important problem in robotic manipulation, yet there still lacks much understanding of how to effectively collect data to facilitate broad generalization. Recent works on large-scale robotic data collection typically vary many environmental factors of variation (e.g., object types, table textures) during data collection, to cover a diverse range of scenarios. However, they do not explicitly account for the possible compositional abilities of policies trained on the data. If robot policies can compose environmental factors from their data to succeed when encountering unseen factor combinations, we can exploit this to avoid collecting data for situations that composition would address. To investigate this possibility, we conduct thorough empirical studies both in simulation and on a real robot that compare data collection strategies and assess whether visual imitation learning policies can compose environmental factors. We find that policies do exhibit composition, although leveraging prior robotic datasets is critical for this on a real robot. We use these insights to propose better in-domain data collection strategies that exploit composition, which can induce better generalization than na\"ive approaches for the same amount of effort during data collection. We further demonstrate that a real robot policy trained on data from such a strategy achieves a success rate of 77.5\% when transferred to entirely new environments that encompass unseen combinations of environmental factors, whereas policies trained using data collected without accounting for environmental variation fail to transfer effectively, with a success rate of only 2.5\%. We provide videos at our project \href{http://iliad.stanford.edu/robot-data-comp/}{website}.\footnote{\url{http://iliad.stanford.edu/robot-data-comp/}}
\end{abstract}

\IEEEpeerreviewmaketitle
\section{Introduction}
\label{sec:introduction}

For robots to be practical and useful, they must be robust to the wide variety of conditions they will encounter in the world. Recent advances in machine learning have shown that leveraging diverse, internet-scale data can be very effective in facilitating this kind of broad generalization \citep{brown2020language, radford2021learning}, as such data captures much of the complexity and variation in the domain that models will need to reason about. It therefore has been of great interest in the robotics community to apply a similar recipe to robotics, specifically using end-to-end imitation learning, which holds promise to effectively scale with large, diverse datasets to achieve broad generalization. However, we lack access to existing internet-scale robotics data, and furthermore, robotics data -- especially human-collected demonstrations -- is often limited and challenging to collect. While recent works have made significant progress in scaling real-world robotic manipulation datasets \citep{walke2023bridgedata, fang2023rh20t, padalkar2023open, gdm2024autort}, their scale is still much less than the datasets typically used for pre-training in computer vision and natural language processing. As a result, policies trained on such datasets often still exhibit unsatisfactory zero-shot performance when deployed in new environments, necessitating additional in-domain data collection for generalization. Given the lack of large-scale robotics data and the high cost of data collection, it is essential to focus on \emph{how to best collect this in-domain data.}

\begin{figure}[t]
    \centering
    \includegraphics[width=\columnwidth]{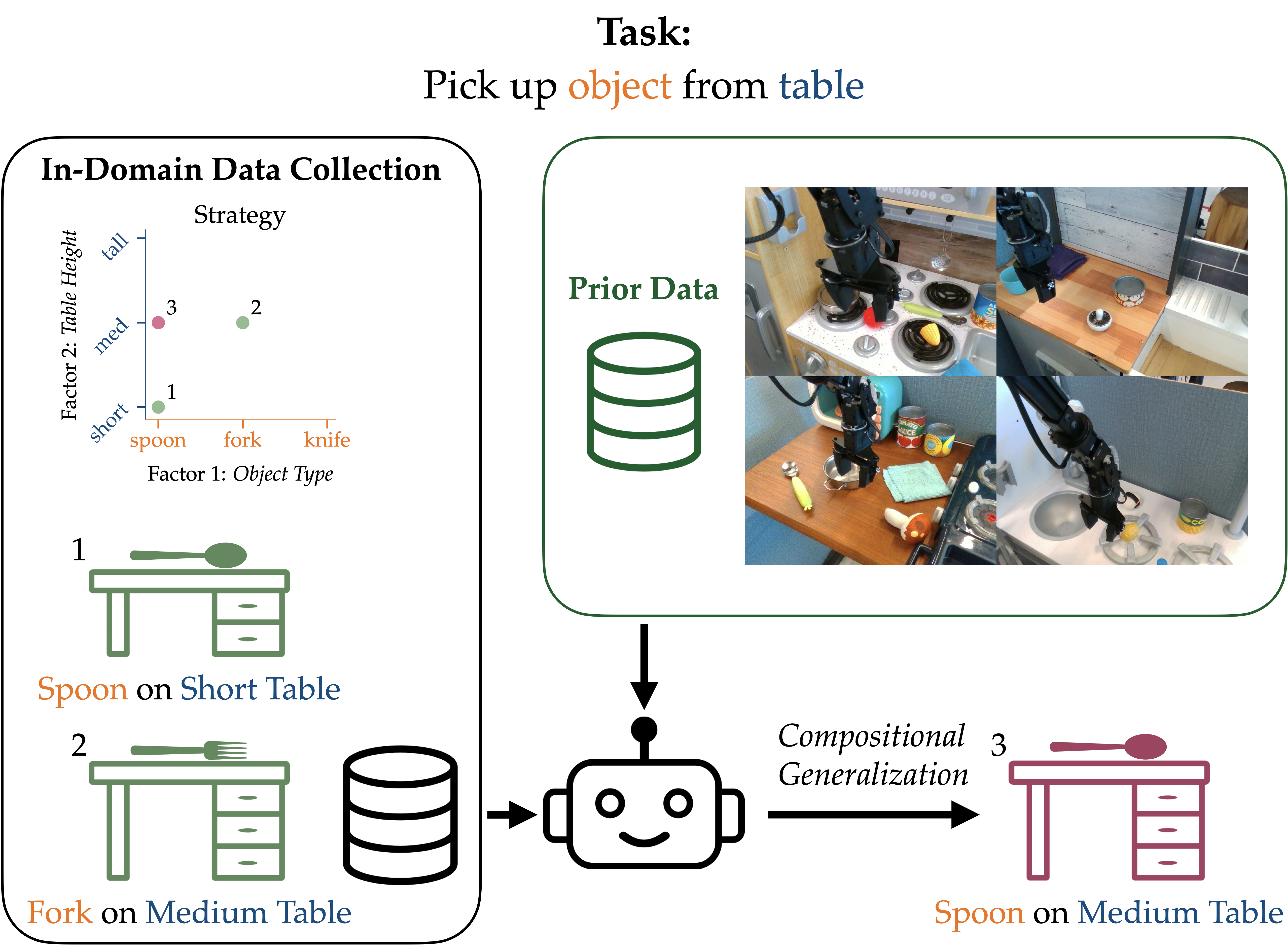}
    \caption{\small{We investigate if robot policies can compose environmental factors of variation (e.g., object types, table heights) in their in-domain training data (1, 2), and if using prior data can be helpful for enforcing such generalization. We propose efficient in-domain data collection strategies guided by the ability of policies to reason about unseen combinations of factor values (3).
    }}
    \label{fig:front}
    \vspace{-20pt}
\end{figure}

When collecting robotics data to account for a given range of scenarios, it is often infeasible to cover all possible settings. Instead, it would be desirable to exploit the \emph{compositional generalization} capabilities of the policies being trained, i.e., their ability to reason effectively about unseen combinations of environmental factors of variation. For example, for the task of picking up different objects from tables of different heights, as shown in~\cref{fig:front}, it would simply not be scalable for humans to collect data for \emph{all} combinations of desired objects and table heights. Instead, it would be much more feasible if one could prioritize collecting data to cover all individual examples of objects and table heights (1 and 2 in \cref{fig:front}), but not necessarily all of their combinations, and then have a policy trained on this data generalize to unseen combinations (3 in \cref{fig:front}). By doing so, we could drastically reduce the amount of data needed.

While prior work has studied generalization to unseen environmental factors in robotics \citep{xing2021kitchenshift, xie2023decomposing, pumacay2024colosseum}, there has not been as much focus on understanding compositional generalization in robotics, particularly for end-to-end imitation learning. Compositional generalization has been previously studied in other machine learning domains \citep{johnson2017clevr, agrawal2017c, lake2018generalization, keysers2019measuring, kim2020cogs, schott2021visual, bogin2021covr, xu2022compositional}, with results often suggesting that end-to-end neural models can struggle to achieve this. However, recent work has shown that large language models can exhibit strong compositional abilities, likely due to the diversity and scale of their training data \citep{zhou2023least, drozdov2023compositional}. Ideally, prior datasets in robotics could also facilitate composition for robotics, but we currently lack much understanding as to where these datasets benefit generalization.

Our key idea is that understanding composition can provide more guidance for efficient data collection in robotics. While there has been a recent trend in scaling up robotic data collection, most prior works attempt to simply collect \emph{diverse} data covering all possible combinations of different factors of variation, without explicitly accounting for composition \citep{walke2023bridgedata, fang2023rh20t, padalkar2023open, gdm2024autort}. However, we hypothesize that imitation learning policies -- similar to their large language model counterparts -- can potentially exhibit compositional generalization, in particular when leveraging prior robotics datasets. In this work, we investigate this hypothesis, with the goal of prescribing better practices for robotic data collection.

Through extensive simulated and real experiments, we study the proficiency of visual imitation learning policies when evaluated on unseen combinations of environmental factors. We consider a wide variety of factors that are broadly relevant in robotic manipulation and have been previously studied, including \emph{object position}, \emph{object type}, \emph{table texture}, \emph{table height}, \emph{camera position}, and \emph{distractor objects}. We find significant evidence of compositional generalization with imitation policies when varying these factors. Given this evidence, we propose in-domain data collection strategies that exploit this generalization -- providing more systematic guidelines that can help reduce data collection burden for roboticists. %
On a real robot, we find that a policy using data from such a strategy succeeds in \textbf{59/90} settings that assess composition, compared to \textbf{22/90} when using the same amount of in-domain data, but without explicit variation. We also find that using prior robot data -- in our case, BridgeData V2 \citep{walke2023bridgedata} -- is critical for this composition, with performance dropping to \textbf{28/90} without it.
Furthermore, we evaluate policies when transferred to entirely new environments that encompass unseen combinations of environmental factors, and find that our best policy achieves a success rate of \textbf{77.5\%}, compared to \textbf{27.5\%} when there is no prior data, and \textbf{2.5\%} when there is no variation in the in-domain data from our training environments.

\vspace{-3pt}
\section{Related Work}
\label{sec:related_work}
In this section, we review prior work on robotic data collection, generalization in robot learning, and compositional generalization more broadly in machine learning.

\smallskip
\noindent \textbf{Data Collection in Robot Learning.}
Prior works have studied data collection methods in robot learning to improve generalization. Much of this work focuses on scaling up robot data collection \citep{pinto2016supersizing, ebert2017self, levine2018learning, mandlekar2018roboturk, sharma2018multiple, dasari2020robonet, ebert2021bridge, brohan2022rt, walke2023bridgedata, fang2023rh20t, gdm2024autort}.
Results from these works have demonstrated that data with greater object and task diversity can improve policy robustness, suggesting that data should be collected with this diversity in mind. In our work, we study how using such datasets as prior data affects compositional generalization. However, results from these works also show that policies trained on these datasets often exhibit poor zero-shot performance in new scenarios, suggesting that in-domain data collection is still often needed in practice. Unlike these prior works, our work focuses on how to collect this in-domain data,
by studying how in-domain data composition
affects generalization at a more nuanced level, and systematically considering a wide variety of specific environmental factors of variation.

Other works on data collection in imitation learning focus on expanding state coverage via interactive imitation learning \citep{ross2011reduction, laskey2017dart, kelly2019hg}, or more recently improving other notions of data quality \citep{belkhale2023data}. These works mainly consider the impact of data on the distribution shift problem \citep{ross2011reduction}, to improve policy robustness to the state distribution the agent will encounter online. However, distribution shift in robotics can also occur due to environmental variations. We instead focus on addressing this form of distribution shift, through offline data collection with active variation of environmental factors.

\smallskip
\noindent \textbf{Generalization in Robot Learning.}
Much prior work has studied improving generalization in robot learning, such as by using pre-trained visual representations \citep{parisi2022unsurprising, nair2022r3m,  ma2022vip, radosavovic2022real, dasari2023unbiased, majumdar2023we, karamcheti2023voltron},
or leveraging diverse robot data \citep{padalkar2023open, octo_2023}.
In our work, we focus on assessing compositional generalization of existing imitation learning methods. Some prior works have similarly studied the robustness of visual imitation learning to environmental factor shifts \citep{xing2021kitchenshift, xie2023decomposing, burns2023makes, pumacay2024colosseum}. However, these works consider generalization to unseen factor values, and primarily focus on how factors individually affect generalization, without extensive focus on how factors interact. Our work instead assesses compositional generalization, where policies must reason about \emph{unseen combinations} of factor values.

\smallskip
\noindent \textbf{Compositional Generalization.}
Compositional generalization has been studied extensively in many machine learning areas, including visual reasoning \citep{schott2021visual, xu2022compositional}, image generation \citep{du2020compositional, liu2022compositional}, natural language processing \citep{lake2018generalization, keysers2019measuring, kim2020cogs}, and visual question answering \citep{johnson2017clevr, agrawal2017c, bogin2021covr}. Much of these works demonstrate that end-to-end neural models often struggle with composition, although sometimes they can exhibit better composition than methods specifically intended to facilitate it, such as unsupervised representation learning \citep{schott2021visual}. However, more recent work has shown that large language models can possess strong compositional abilities, likely owing to their internet-scale pre-training data \citep{zhou2023least, drozdov2023compositional}. In our work, we study if leveraging prior data can also be beneficial for composition in robotics.

Compositional generalization has also been investigated in robotics. This includes work on modular approaches for promoting composition \citep{devin2017learning, xu2018neural, kuo2020deep, wang2023programmatically}, and end-to-end architectures with inductive biases that benefit composition \citep{zhou2022policy}.
However, these works primarily study composition at the semantic level, for understanding concepts such as text instructions, high-level skills, or object properties and relationships. Unlike these works, we study composition of \emph{environmental factors}, which may significantly change visual observations, or the low-level physical motions needed for a task. In addition, we study composition for existing end-to-end imitation learning methods. Such approaches have become popular in robotics and hold promise to scale effectively with data. %

There has been some evidence of compositional generalization in end-to-end robotic imitation learning. This includes composition of tasks in prior data with the conditions of a target domain \citep{ebert2021bridge}, objects with manipulation skills \citep{brohan2022rt, stone2023open}, semantic concepts from internet data with manipulation skills \citep{rt22023arxiv}, and behaviors and semantic concepts across robot embodiments \citep{brohan2022rt, padalkar2023open}. However, our work provides more systematic and fine-grained analysis on when composition is possible, and considers a greater variety of environmental factors. Furthermore, unlike these works, we focus on using this analysis to guide better practices for data collection that take composition into account. We believe these key differences allow our insights to be leveraged for more effective data collection in real-world settings.

\section{Exploiting Compositional Generalization for Efficient Data Collection}
\label{sec:motivation}

\vspace{5pt}
\subsection{Problem Statement}
We consider the goal-conditioned imitation learning setting, where the objective is for a robot to reach a goal. We formalize this by defining the tuple $(\mathcal{S}, \mathcal{A}, \mathcal{T}, H, \mathcal{F}^N, p(f), \rho(s_0 | f), \mu(\cdot | f))$, where $\mathcal{S}$ and $\mathcal{A}$ are the state and action spaces, $\mathcal{T}(s'|s, a)$ is the transition dynamics function, and $H$ is the horizon length. We refer to $\mathcal{F}^N \subset \mathbb{Z}^N$ as the \emph{factor space}, which captures changes in the environment along $N$ different axes that the robot could encounter, which we refer to as \emph{factors}. For example, the first component of $\mathcal{F}^N$ could capture variation in \emph{object type}, and the second could capture variation in \emph{table height}, etc. For simplicity, we assume each component of $\mathcal{F}^N$ consists of $k$ possible discrete values, e.g., there are $k$ possible object types the robot may encounter. We refer to these as \emph{factor values}. Therefore, each element $f \in \mathcal{F}^N$ defines a combination of factor values for the environment, e.g., a specific \emph{object type} and \emph{table height} the robot may encounter. $p(f)$ is the distribution of factor value combinations, and $\rho(s_0 | f)$ is the initial state distribution, conditioned on a combination of factor values. We adopt this formalism to express that factor values in the environment can affect tasks by inducing different initial states. Similarly, $\mu(f) : \mathcal{F}^N \rightarrow S$ is a deterministic function that maps a combination of factor values to the desired goal state $g \in S$, as the factor values can determine what the goal state is for a task. We aim to learn a goal-conditioned policy $\pi(\cdot|s, g): S \times S \rightarrow \Delta(\mathcal{A})$, where $\Delta(\mathcal{A})$ is the probability simplex over the action space. We aim to maximize the probability of reaching the desired goal, under the distribution of possible factor values:
\begin{equation}
    J(\pi) = \mathbb{E}_{f \sim p(f)} [P_\pi (s_H = \mu(f))],
\end{equation}
where $P_\pi (s_H = \mu(f))$ is the probability that the state visited by policy $\pi$ at the final timestep $H$ is the goal state $g = \mu(f)$.

To learn the policy $\pi$, we assume a dataset of $M$ expert demonstrations $\mathcal{D}_M = (\tau_1, \dots, \tau_M)$. Each demonstration $\tau_i = \{(s_1, a_1), \dots, (s_{T_i}, a_{T_i})\}$ is a sequence of state-action pairs of length $T_i$, produced by sampling actions from an expert policy $\pi_E(\cdot|s, g)$ through dynamics $\mathcal{T}(s'|s, a)$. Each demonstration also has an associated combination of factor values $f_i$, which defines both the initial state distribution the demonstration was collected from, and the goal state $g = \mu(f)$ the expert policy aimed to reach. We can then use this dataset to learn a policy $\pi$ using goal-conditioned behavior cloning \citep{lynch2020learning}.

\subsection{Objective}
We aim to train robotic manipulation policies using imitation learning that will robustly handle combinations of factor values from some desired distribution $p(f)$, by collecting demonstrations with associated factor values $f$ according to some strategy.  We assume without loss of generality that $p(f) = \text{Uniform}(\mathcal{F}^N)$, such that we desire to handle all factor value combinations in $\mathcal{F}^N$. The most direct way to achieve this is sufficiently covering all of $\mathcal{F}^N$, such as by sampling new values $f \sim \text{Uniform}(\mathcal{F}^N)$ enough times, or even by collecting demonstrations for all $f \in \mathcal{F}^N$. However, as the cardinality of $\mathcal{F}^N$ increases exponentially with the number of factors $N$, collecting demonstrations for all possible factor value combinations is often impractical. Furthermore, constantly resampling new values $f \sim \text{Uniform}(\mathcal{F}^N)$ can often be expensive or challenging in practice, as each factor change often requires some degree of manual effort.

Rather than attempting to collect data for all factor value combinations in $\mathcal{F}^N$, it would be desirable to instead collect data with a focus on capturing \emph{individual} factor values, and then exploit \emph{compositional generalization} of the learned policy to perform well on unseen combinations. With $N$ factors and $k$ possible values each, collecting data for all combinations would require $\mathcal{O}(k^N)$ factor changes in the environment, while covering all individual factor values would only require $\mathcal{O}(kN)$ changes, a dramatic reduction. However, this is a sensible option only if composition is actually achieved by the learned policy. As prior work has shown mixed results on the compositional abilities of end-to-end neural models, it remains important to investigate \emph{when} compositional generalization happens in imitation learning for robotic manipulation. Therefore, to exploit compositional generalization for more efficient data collection, we must answer the following questions:

\begin{enumerate}
    \item When do robotic imitation learning policies exhibit compositional generalization?
    \item What are effective data collection strategies to exploit composition, such that we can achieve broad generalization while reducing data collection effort?
\end{enumerate}

\noindent We answer these questions by conducting extensive experiments in both simulation and on a real robot, investigating the compositional abilities of imitation learning policies for a variety of factors. We compare data collection strategies intended to exploit composition by optimizing for coverage of individual factor values, as opposed to na\"ively optimizing for coverage of all factor combinations.

\begin{figure}[ht]
    \centering
    \includegraphics[width=\columnwidth]{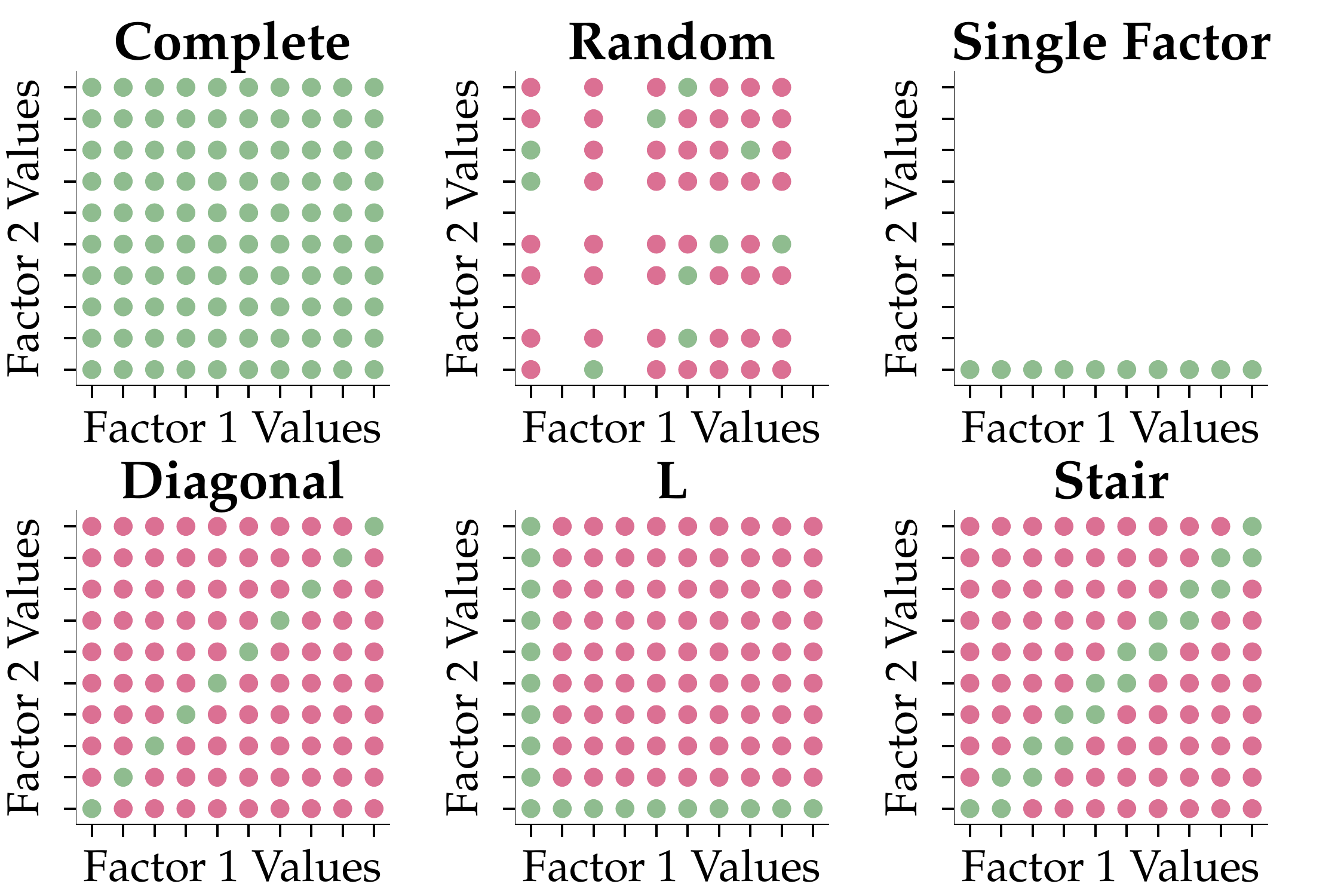}
    \vspace{-15pt}
    \caption{\small{Visualization of our data collection strategies with $N = 2$ factors. Each axis consists of possible values for a factor. Each green dot indicates that the strategy captures a specific combination of factor values represented by it, and each pink dot represents a combination that compositional generalization may address. We name our strategies based on the patterns in this visualization.}}
    \label{fig:data_strats}
    \vspace{-20pt}
\end{figure}

\vspace{5pt}
\subsection{Data Collection Strategies}
\label{sec:data_strats}
We describe the data collection strategies that we consider. We visualize them for when the number of factors $N = 2$ in \cref{fig:data_strats}, but these can easily be extended for $N > 2$. We also provide pseudocode for \textbf{Diagonal}, \textbf{L}, and \textbf{Stair} in \cref{sec:strat_pseudocode}. The strategies involve periodically setting factor values in the environment during data collection. To quantify effort, we consider the total number of factor changes made in the environment. Note that changing multiple factors at a time involves multiple changes, e.g., setting the environment to have different values for all factors involves $N$ changes.

\begin{enumerate}
    \item \textbf{Complete}: This strategy covers all combinations of factor values in $\mathcal{F}^N$. As this requires $\mathcal{O}(k^N)$ factor changes, this is often deemed infeasible in practice.
    \item \textbf{Random}: This strategy periodically resamples an entire random combination of factor values $f$ from all possible $f \in \mathcal{F}^N$ without replacement. This strategy will eventually cover all of $\mathcal{F}^N$, but may achieve generalization inefficiently by not actively leveraging composition.
    \item \textbf{Single Factor}: This strategy only varies one factor during data collection, while keeping the values of all other factors the same as some base factor values $f^*$.
    \item \textbf{Diagonal}: This strategy resamples entire combinations of factor values $f$, but only samples $f$ where each factor value has never been seen. This requires $\mathcal{O}(kN)$ factor changes to cover all values, the fewest possible.
    \item \textbf{L}: This strategy varies only one factor at a time. Starting from base factor values $f^*$, the possible combinations of factor values $f$ only deviate from $f^*$ by one factor. This also covers all factor values with $\mathcal{O}(kN)$ factor changes.
    \item \textbf{Stair}: This strategy starts at $f^*$, and then cyclically varies one factor at a time, while preserving the values for all other factors. This also covers all factor values with $\mathcal{O}(kN)$ changes. However, compared to \textbf{Diagonal}, it covers more combinations of factor values, and compared to \textbf{L}, it captures more diverse combinations.
    \item \textbf{No Variation}: This baseline involves only collecting data for a single base combination of factor values $f^*$.
\end{enumerate}

\noindent \textbf{Stair}, \textbf{L}, and \textbf{Diagonal} are intended to exploit compositional generalization by prioritizing covering  individual factor values with the fewest amount of factor changes. We note that using factor changes as a measure of effort is a rough approximation. For example, this may not be accurate when it is more challenging to vary two factors together than separately. This could be the case if one had to collect data in multiple separate environments to vary a factor like \emph{table texture}, but could collect data in a single environment to vary \emph{object type}. In such situations, it may be easier to use the \textbf{L} strategy to collect data varying \emph{object type} in only one environment, and collect data varying \emph{table texture} in separate environments without varying \emph{object type}, instead of varying both factors together like \textbf{Stair} and \textbf{Diagonal} would require.

\vspace{-3.5pt}
\subsection{Hypotheses}
\noindent We develop the following hypotheses, which we seek to investigate in our experiments:
\begin{enumerate}
    \item \textbf{Stair}, \textbf{L}, and \textbf{Diagonal} may outperform \textbf{Random} by exploiting compositional generalization when possible.
    \item \textbf{Stair}, \textbf{L}, and \textbf{Diagonal} may approach the performance of \textbf{Complete} when composition is strong.
    \item \textbf{Stair} may outperform \textbf{L} and \textbf{Diagonal}, by capturing a greater quantity and diversity of factor combinations.
    \item Incorporating prior robot data can promote stronger composition of factor values in the in-domain data.
\end{enumerate}

\vspace{-5pt}
\section{Simulation Experiments}
\label{sec:sim_experiments}

To evaluate composition and data collection strategies at a large scale, we first conduct extensive experiments in simulation. We use \emph{Factor World}, a robotics simulation platform that supports varying different environmental factors \citep{xie2023decomposing}.

\begin{figure}[h!]
    \vspace{-3pt}
    \centering
    \includegraphics[width=\columnwidth]{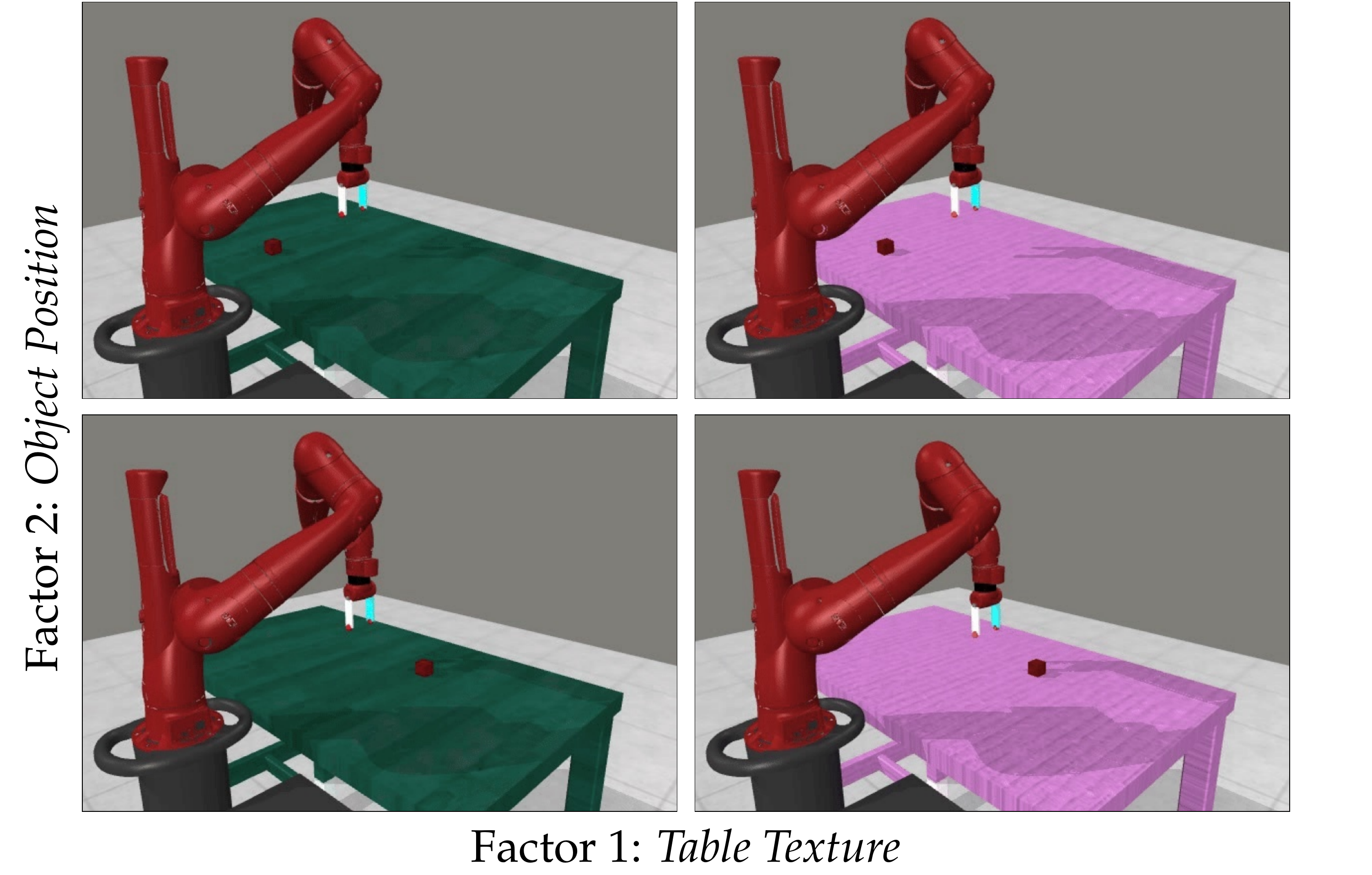}
    \vspace{-15pt}
    \caption{\small{Visualization of the \emph{Pick Place} task. We show combinations of 2 values each for the factors \emph{table texture} and \emph{object position}.}}
    \label{fig:sim_ex}
    \vspace{-12pt}
\end{figure}

\begin{figure*}[t]
    \centering
    \includegraphics[width=\textwidth]{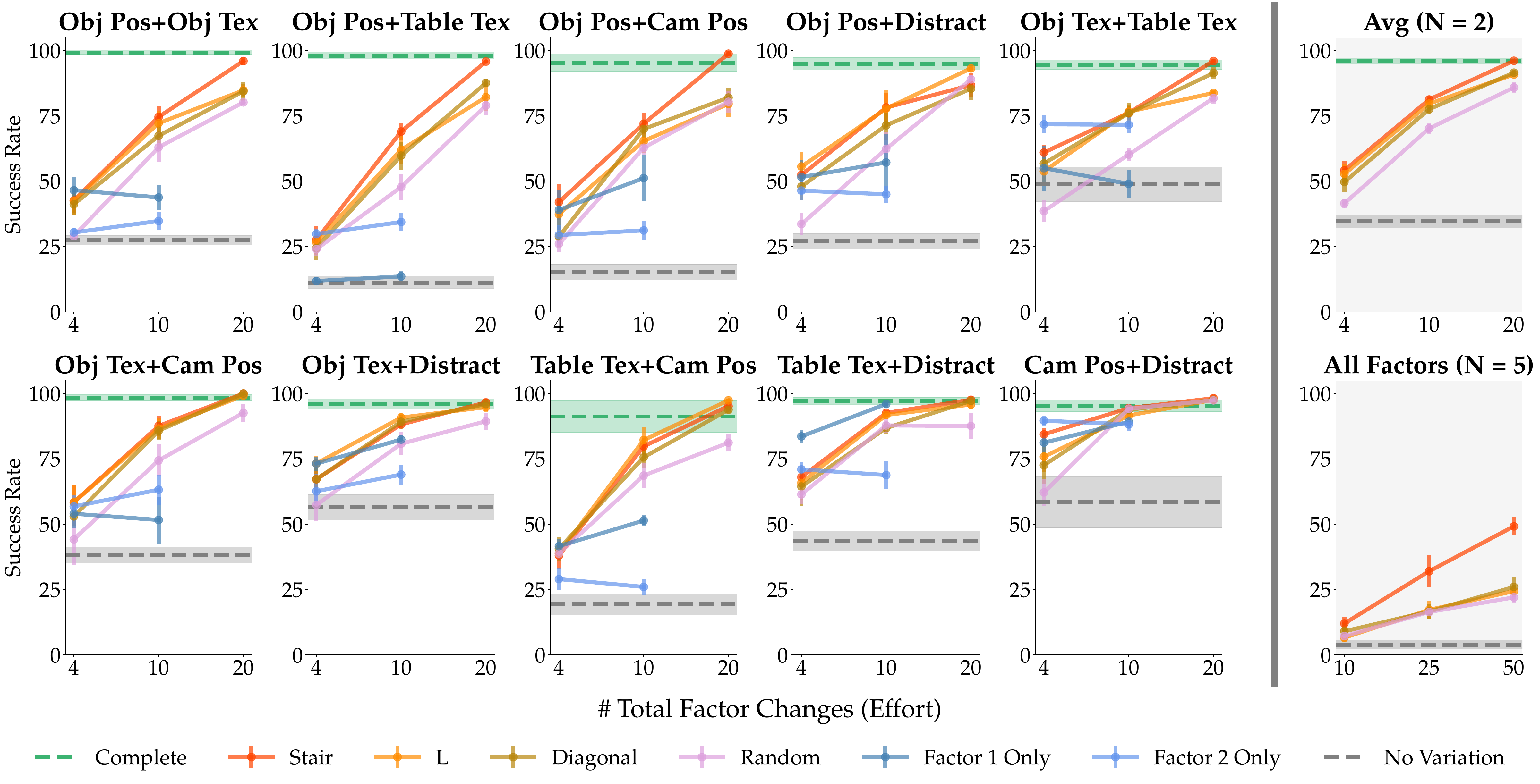}
    \vspace{-17pt}
    \caption{
    \small{Simulation results of data collection strategies for \emph{Pick Place}. We report results where $\mathcal{F}^N$ consists of each possible factor pair ($N = 2$), average results across all pairs, and results where $\mathcal{F}^N$ consists of all factors ($N = 5$). All points within the same subplot use the same amount of demonstrations. The strategies that exploit composition (\textbf{Stair}, \textbf{L}, \textbf{Diagonal}) generally outperform \textbf{Random}, and often approach \textbf{Complete}. \textbf{Stair} generally performs the best, especially in the $N = 5$ setting. Error bars represent standard error across 5 seeds.}}
    \label{fig:pick_place_results}
    \vspace{-17.5pt}
\end{figure*}

\vspace{-1pt}
\subsection{Evaluation Protocol}
We consider \emph{Pick Place} and \emph{Door Open}, two commonly studied tasks in robotics. We consider 5 factors, including 4 that generalization was challenging for in the original \emph{Factor World} experiments (\emph{object texture}, \emph{table texture}, \emph{camera position}, and \emph{distractor objects}), as well as \emph{object position}, as an example of a factor that more strongly affects required physical motion. We show example combinations of \emph{table texture} and \emph{object position} for \emph{Pick Place} in \cref{fig:sim_ex}, and examples of individual factor values for both tasks in \cref{sec:sim_eval_details}. To study which specific factors can be composed by policies effectively, we consider the settings where $\mathcal{F}^N$ consists of one of the 10 possible factor pairs ($N = 2$). To study composition in a more realistic setting where robustness to more factors is desired, we also consider when $\mathcal{F}^N$ consists of all factors ($N = 5$). For each factor, we sample $k = 10$ values that the factor can take in $\mathcal{F}^N$. For each data strategy, we use a scripted expert policy to collect multiple datasets with different total amounts of factor changes, by setting new factor values according to the strategy at different rates. For example, 20 factor changes would involve setting new values roughly twice as often as with 10 changes. When setting new values, we choose which values to set at random from what is permissible by the current strategy. We compare how the strategies scale with the total number of factor changes, which we use to quantify \emph{effort}.

We only compare against \textbf{Complete} in the $N = 2$ setting, as $N = 5$ would require a minimum of $10^5$ demonstrations, which is impractical. Each dataset consists of 100 demonstrations for the $N = 2$ experiments, and 1000 demonstrations for $N = 5$. We train a policy on each dataset using behavior cloning, and evaluate on 100 episodes, each with a different $f$ randomly sampled from $\mathcal{F}^N$, to assess overall robustness to factor combinations. For the $N = 2$ experiments, this consists of all possible $f \in \mathcal{F}^N$. We evaluate across 5 random seeds, which include different possible factor values in $\mathcal{F}^N$ for each seed. We provide more details in \cref{sec:sim_eval_details}.

\subsection{Results}
\label{sec:sim_results}
\vspace{-5pt}
We illustrate our results for \emph{Pick Place} in \cref{fig:pick_place_results}. We first focus on the $N = 2$ setting. We see that \textbf{No Variation} (dashed gray) performs poorly, verifying that a variety of factor values is needed to generalize.  The \textbf{Single Factor} strategies (shades of blue) also perform poorly for most factor pairs, verifying that most pairs require variation in both factors to generalize. Note that these strategies can vary only a single factor at most 10 times, leading to shorter lines in the plots. We see that the strategies intended to exploit composition -- \textbf{Stair}, \textbf{L}, \textbf{Diagonal} (shades of orange) -- generally outperform \textbf{Random} (light purple) at all levels of factor changes. They also approach the performance of \textbf{Complete} (dashed green line) with only 20 factor changes (enough to capture all individual factor values), compared to the 100 changes needed for \textbf{Complete}. These results suggest that these policies exhibit strong pairwise compositional abilities for these factors, and that \textbf{Stair}, \textbf{L}, and \textbf{Diagonal} are able to exploit this for more efficient data collection. We also see that \textbf{Stair} does slightly better overall than \textbf{L} and \textbf{Diagonal}, possibly suggesting some benefit from a greater quantity and diversity of factor value combinations.

\begin{figure}[t]
    \centering
    \includegraphics[width=\columnwidth]{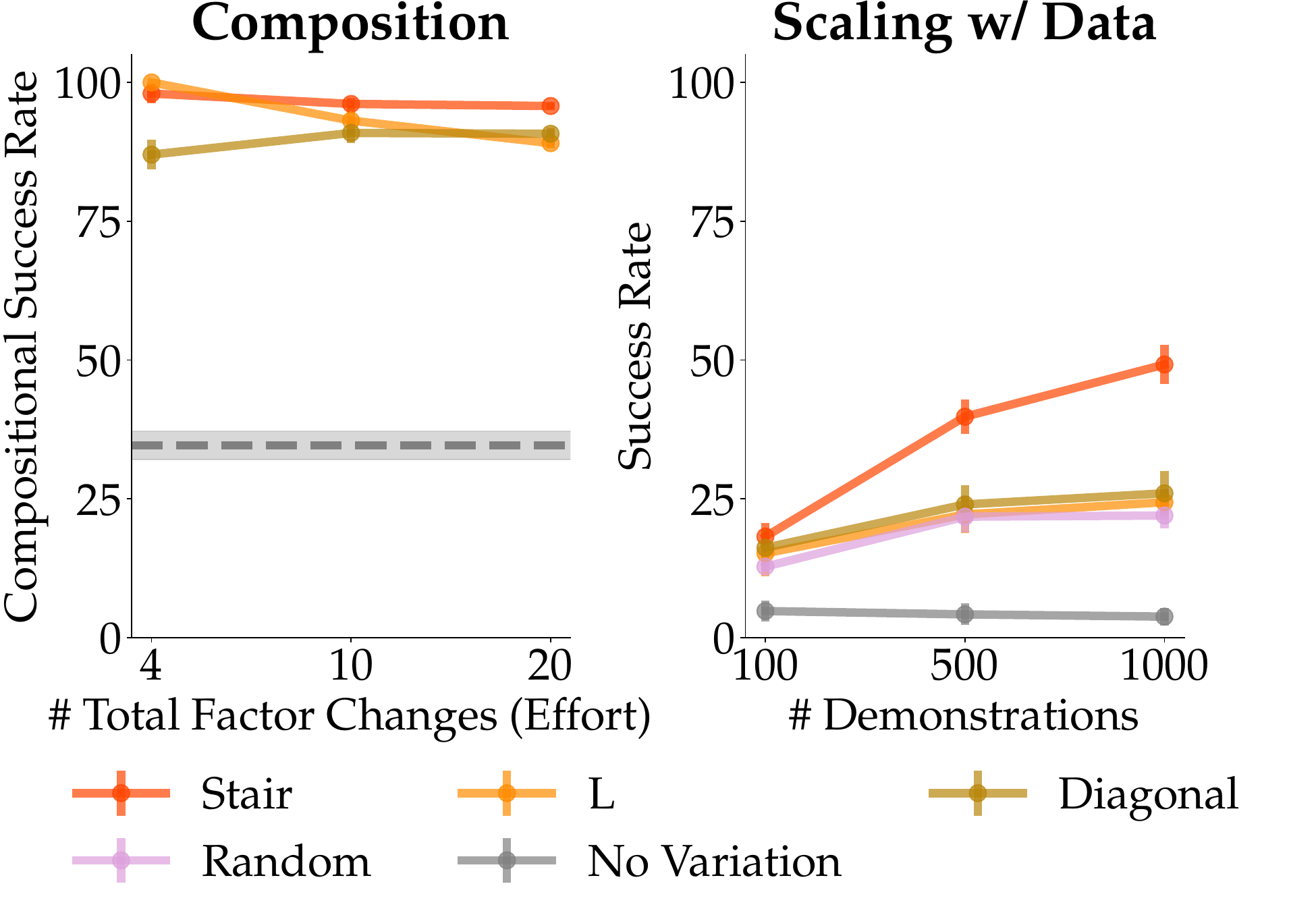}
    \vspace{-15pt}
    \caption{\small{(left) Compositional success rate of different strategies.\\(right) Generalization of strategies with increasing dataset sizes.}}
    \label{fig:additional_results}
    \vspace{-20pt}
\end{figure}

In the $N = 5$ setting, performance is lower overall, indicating that generalization is harder when more factors are varied. Furthermore, \textbf{Stair} is now the only strategy that substantially improves over \textbf{Random}. This suggests that a greater quantity and diversity of factor value combinations is more important for composition when there are more factors involved.

We provide similar results for \emph{Door Open}, results with different data augmentation, and results on accounting for factor value similarity during data collection in \cref{sec:sim_eval_details}.

\smallskip \noindent \textbf{Assessing Composition.} In \cref{fig:additional_results} (left), we more closely investigate composition for the strategies intended to exploit it (\textbf{Stair}, \textbf{L}, \textbf{Diagonal}), and how this scales with the number of factor values seen. Unlike before, here we report each policy's \emph{compositional success rate}: the success rate \textbf{only} on combinations of factor values where each factor value was seen individually, but not in the exact same combination. The previous results considered all possible combinations, to assess overall generalization. We report this metric averaged across all pairs from our $N = 2$ evaluation. We see these strategies exhibit strong composition for all levels of factor changes, suggesting that composition does not require observing a large number of factor values. However, \textbf{Stair} does appear to induce composition slightly better than the other strategies.

\smallskip \noindent \textbf{Scaling with Dataset Size.} In \cref{fig:additional_results} (right), we investigate how generalization scales with dataset size. We focus on the $N = 5$ setting, where generalization is more challenging, and therefore the impact of data scale is more pronounced. We report the success rate of policies trained on datasets from different strategies, which all have 50 factor changes (the amount needed for \textbf{Stair}, \textbf{L}, and \textbf{Diagonal} to capture all factor values), but different amounts of demonstrations. We also compare against \textbf{No Variation}, which does not scale well, verifying that simply more data is not enough to generalize. The other strategies do scale positively, but \textbf{Stair} does significantly better than the rest, and it is the \textbf{only strategy} that is able to significantly benefit when increasing from 500 to 1000 demonstrations. This suggests that having more data, even without greater factor diversity, can help facilitate improved compositional generalization, but having a greater quantity and diversity of factor combinations, which \textbf{Stair} provides, may be critical for enabling this scaling.

\vspace{-3pt}
\section{Real Robot Experiments}
\label{sec:real_experiments}
Our simulation results have been encouraging regarding composition in robotic imitation learning, and the potential to exploit this with strategies for more efficient data collection. However, for such strategies to be useful for real-world data collection, real robot policies must also have compositional abilities. Therefore, we conduct experiments to evaluate if and when compositional generalization happens on a real robot. In addition, we include experiments incorporating prior robotic datasets, to investigate if access to such datasets can benefit composition in robotics, as is the case with natural language.

\vspace{-5pt}
\subsection{Evaluation Protocol}
\begin{figure}[t]
    \centering
    \includegraphics[width=\columnwidth]{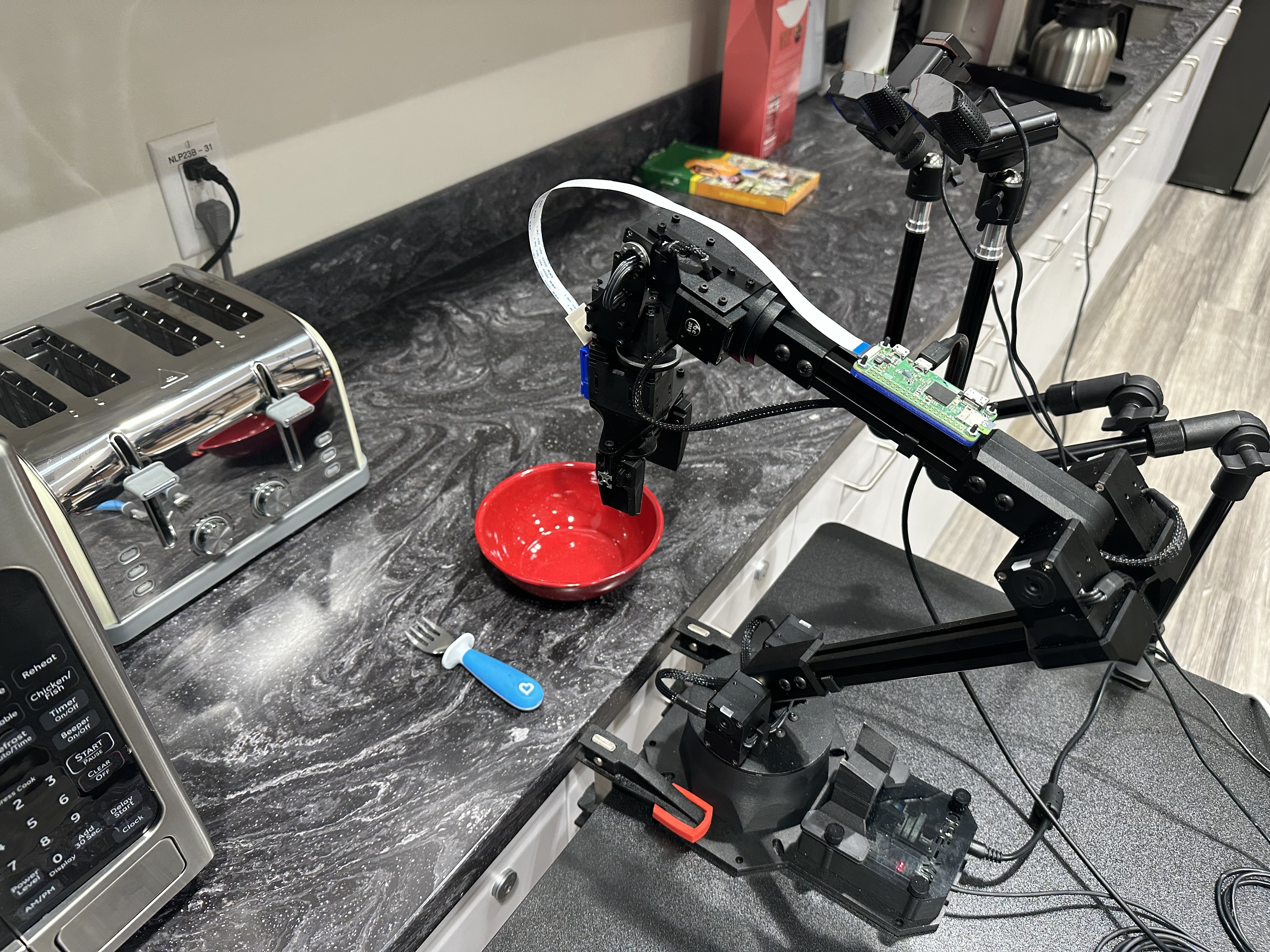}
    \caption{\small{Our WidowX robot setup. We consider the task of putting a fork inside a container, in a real office kitchen.}}
    \label{fig:robot_setup}
    \vspace{-20pt}
\end{figure}

\noindent \textbf{Robot Platform.}
We conduct experiments using a WidowX 250 6DOF robot arm. We adapt the hardware and control setup used in BridgeData V2 \citep{walke2023bridgedata}, except we directly mount the robot and over-the-shoulder RGB camera to a mobile table. While our hardware setup is not identical to the original, we still seek to leverage BridgeData V2 as prior data, which represents a realistic use case of a prior robotic dataset. To promote transfer, we tune our camera setup so that policies trained on only BridgeData V2 can sometimes perform basic pick-place tasks zero-shot, although we were unable to have it perform the exact tasks in our evaluation. We show our robot setup in \cref{fig:robot_setup}, and provide more details in \cref{sec:robot_platform_details}.

\begin{figure*}[t]
    \centering
    \includegraphics[width=\textwidth]{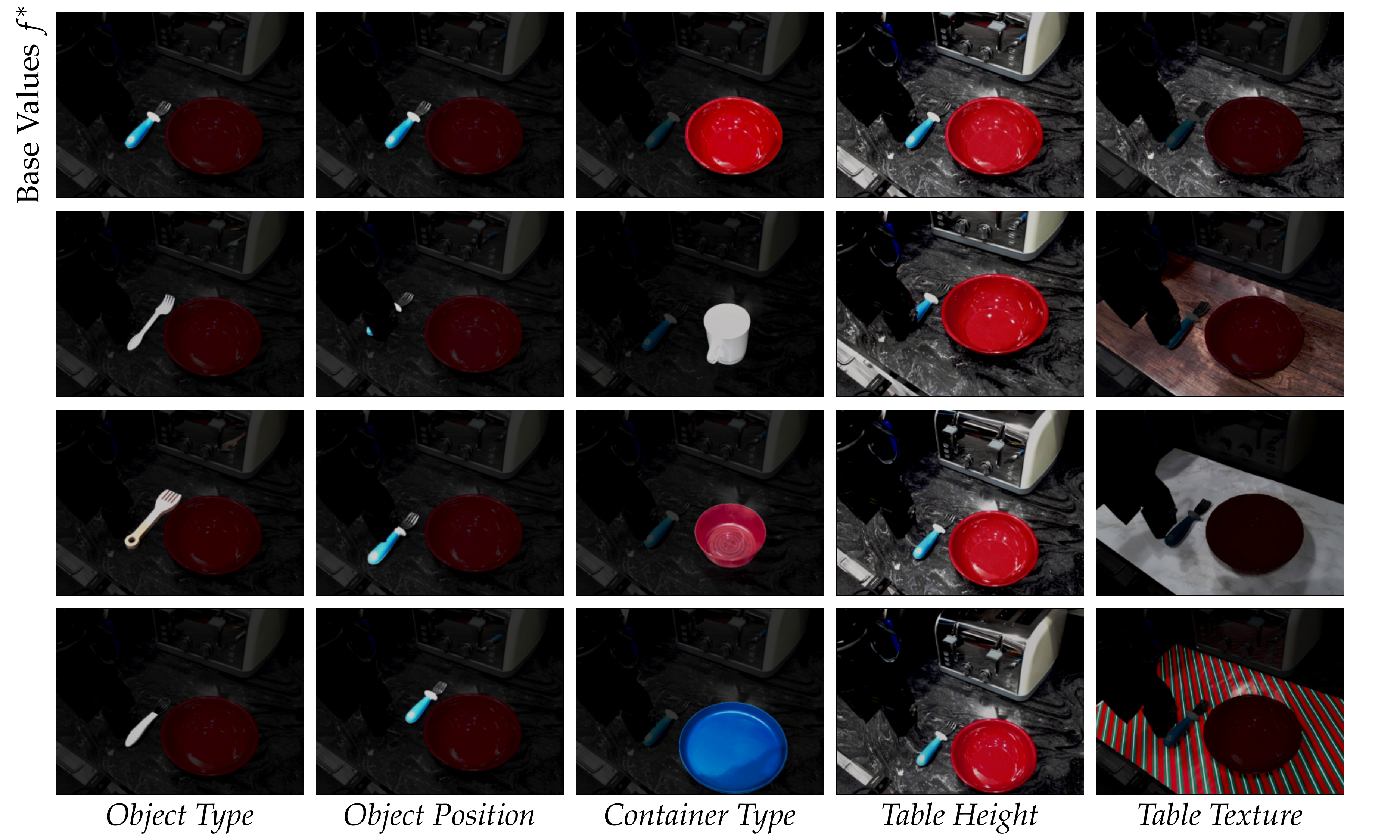}
    \vspace{-15pt}
    \caption{\small{Visualization of our primary real robot factors in \emph{BaseKitch}. The top row shows our base factor values $f^*$. The other rows show all deviations from $f^*$ by one factor value. We highlight the portion of each image affected by the specified factor value.}}
    \label{fig:real_factors}
    \vspace{-17.5pt}
\end{figure*}

\smallskip \noindent \textbf{Task and Factors.} We consider the task of placing a fork inside a container, on the countertop of an office kitchen. We primarily consider the following factors: \emph{object (fork) type}, \emph{object (fork) position}, \emph{container type}, \emph{table height}, and \emph{table texture}. Each factor has $k = 4$ possible values. We visualize our base values $f^*$, and also every deviation from $f^*$ by one factor, in \cref{fig:real_factors}. We additionally consider the factors \emph{object/container position}, \emph{object (fork) orientation}, \emph{distractor objects}, and \emph{camera position} in secondary experiments.

Compared to our simulation experiments, our real experiments have a greater emphasis on \emph{physical} factors that affect required physical motions, as opposed to \emph{visual} factors that a policy should be mostly invariant to. For instance, we only study \emph{object type}, \emph{container type}, and \emph{table height} in real, which are \emph{physical} factors that can affect required grasp and place motions. This is because it is more difficult to conduct large-scale simulation experiments for such factors, due to the challenge of automatically collecting demonstrations that account for them. We believe that addressing \emph{physical} factors is especially important because they are likely more resistant to visual data augmentation, and thus may require more emphasis during data collection. Also, \emph{physical} factors are less likely to benefit from using internet data for semantic reasoning \citep{ichter2022do, rt22023arxiv}.

\smallskip \noindent \textbf{Experimental Setup.}
We adopt a different experimental setup for our real experiments, due to the much greater cost of collecting data, training policies, and evaluating. We collect datasets for only the $N = 5$ setting with all of our primary factors. Because the objective of these experiments is to determine which factors can be composed by a real robot policy, we focus on data collection strategies intended to exploit composition. In particular, we evaluate \textbf{Stair}, because it performed the best in simulation, and \textbf{L}, because by collecting data for each factor separately, we can use this to study pairwise composition using a model trained on a single dataset. \textbf{L} may also be more practical in real-world situations where it is easier to collect data for factors separately. For these strategies, we collect 10 human demonstrations for each of the 16 combinations of factor values required to cover all values per factor, resulting in 160 total demonstrations for each strategy. We also compare against \textbf{No Variation} with the same number of demonstrations. We collect data exclusively in one kitchen, which we refer to as \emph{BaseKitch}. We train policies either from scratch, or with the addition of BridgeData V2, to assess the impact of prior data on composition. We provide more details on our experimental setup in \cref{sec:real_eval_details}.

\begin{table*}[t]
    \begin{adjustbox}{width=\textwidth}
    \begin{tabular}{l|cccc|cccc|cccc}
        \toprule
        Data Strategy   & \multicolumn{8}{c|}{\textbf{L}}                                                                                        & \multicolumn{4}{c}{\textbf{No Variation}} \\
        \midrule
        Train Method    & \multicolumn{4}{c|}{Bridge}                                                    & \multicolumn{4}{c|}{From Scratch}                    & \multicolumn{4}{c}{Bridge} \\
        \midrule
        \diagbox{Factor 1}{Factor 2}
                        & \makecell{Object\\Type} & \makecell{Container\\Type} & \makecell{Table\\Height} & \multicolumn{1}{l|}{\makecell{Table\\Tex}}
                        & \makecell{Object\\Type} & \makecell{Container\\Type} & \makecell{Table\\Height} & \multicolumn{1}{l|}{\makecell{Table\\Tex}}
                        & \makecell{Object\\Type} & \makecell{Container\\Type} & \makecell{Table\\Height} & \makecell{Table\\Tex} \\
        \midrule
        Object Pos      & \textbf{8/9} & \textbf{5/9} & \textbf{2/9} & \multicolumn{1}{c|}{\textbf{5/9}} & 0/9 & 0/9  & 0/9 & \multicolumn{1}{c|}{1/9}          & 3/9 & 2/9 & 1/9 & 1/9      \\
        Object Type     &              & \textbf{8/9} & \textbf{8/9} & \multicolumn{1}{c|}{\textbf{8/9}} &     & 5/9  & 5/9 & \multicolumn{1}{c|}{4/9}          &     & 4/9 & 3/9 & 2/9      \\
        Container Type  &              &              & \textbf{5/9} & \multicolumn{1}{c|}{\textbf{6/9}} &     &      & 4/9 & \multicolumn{1}{c|}{\textbf{6/9}} &     &     & 2/9 & 3/9      \\
        Table Height    &              &              &              & \multicolumn{1}{c|}{\textbf{4/9}} &     &      &     & \multicolumn{1}{c|}{3/9}          &     &     &     & 1/9      \\
        \midrule
        Overall                        &  \multicolumn{4}{c|}{\textbf{59/90}}                            & \multicolumn{4}{c|}{28/90}                           & \multicolumn{4}{c}{22/90}  \\
        \bottomrule
    \end{tabular}
    \end{adjustbox}
    \caption{\small{Real robot pairwise composition results for our ``\emph{put fork in container}" task}. When leveraging BridgeData V2 as prior data, a policy is able to compose the factor values present in the \textbf{L} data to succeed on \textbf{59/90} compositional combinations of factor values. Without prior data, the model is unable to compose nearly as effectively, with compositional success rate dropping by roughly half. Prior data alone is also not enough to generalize to these situations, as a policy trained with prior data on \textbf{No Variation} also performs poorly.}
    \label{tab:pairwise_results}
    \vspace{-19pt}
\end{table*}

\smallskip \noindent \textbf{Training.}
We primarily train policies based on the diffusion goal-conditioned behavior cloning method proposed in BridgeData V2 \citep{walke2023bridgedata}, adapting the implementation provided 
in their code. For policies that use BridgeData V2 as prior data, unless otherwise stated, we first pre-train a policy on only the prior data, and then co-fine-tune on a mixture of in-domain and prior data. We verify that all policies succeed with base factor values $f^*$ in \emph{BaseKitch}. We provide more details in \cref{sec:real_train_details}.

\vspace{-2pt}
\subsection{Pairwise Composition}
\label{sec:pairwise}
In this evaluation, we study pairwise composition of factors in \emph{BaseKitch}. We compare \textbf{L} and \textbf{No Variation} to evaluate composition. For each pair of factors, we evaluate on all compositional combinations of factor values for that pair with respect to the \textbf{L} dataset, and set the values for other factors to their base values in $f^*$. This results in 9 evaluation scenarios for each of the 10 pairs.\footnote{We have 3 values per factor that are not in the base values $f^*$, resulting in $3^2=9$ combinations for each of the $\binom{5}{2}=10$ pairs.} For \textbf{L}, we compare both training from scratch, and using BridgeData V2 as prior data. For \textbf{No Variation}, we only evaluate training with BridgeData V2.

We report our results in \cref{tab:pairwise_results}. We find that by training on \textbf{L} data and using BridgeData V2, our policy is able to generalize to \textbf{59/90} of the compositional factor value combinations. This suggests that composition is possible in real-world robotic imitation learning, similar to our simulation results. However, unlike in simulation, leveraging prior data is critical for strengthening composition. Without prior data, the policy's overall compositional success rate is roughly halved to \textbf{28/90}, with success rates dropping for \textbf{9/10} factor pairs. This resembles results in prior work that suggest end-to-end neural models can struggle with composition, but large pre-trained models can exhibit strong composition \citep{zhou2023least, drozdov2023compositional}. We hypothesize that prior data may be needed in real, but not simulation, due to additional minor factor variations inherent to real experiments, including those for factors that we do not account for. Our greater emphasis on \emph{physical} factors in real may also account for some of this discrepancy.

However, prior data alone is not enough for generalization, as training on \textbf{No Variation} with prior data performs much worse than using \textbf{L} data, achieving a success rate of only \textbf{22/90}. This suggests that with current prior robotic datasets, varied in-domain data is often still critical for generalization, motivating the need for efficient in-domain data collection.

We note that composition is generally strongest for pairs where at least one factor is \emph{visual}. We hypothesize this is because \emph{physical} factors can interact in more complex ways, making it more challenging for policies to compose unseen combinations of them. For example, \emph{object position} and \emph{table height} have the weakest composition together, possibly because both significantly affect the required grasp motion, and thus composition requires executing completely unseen grasps. Similarly, composition for \emph{container type} is the weakest for the value \textbf{white cup} (second row in \cref{fig:real_factors}), which is narrower and taller than the other containers, and therefore requires a different place motion. In particular, this composes poorly with \emph{object position} and \emph{table height}, both \emph{physical} factors.

In contrast, composition is generally the strongest for \emph{object type}. Although \emph{object type} may affect grasp and place motions, especially when interacting with other \emph{physical} factors, we hypothesize it composes well because the different forks we consider are similar enough to not require drastically different motions. Overall, these results suggest that even with current prior robotics datasets, composition between \emph{physical} factors can still be challenging, and thus more exhaustive coverage for these factors during data collection may be necessary.

We provide further pairwise composition results for additional factors \emph{object/container position} and \emph{object orientation} in \cref{sec:additional_pairwise}. In \cref{sec:similarity_analysis}, we provide additional analysis on how pairwise composition is affected by how similar new factor values are to the base factor values $f^*$.

\smallskip \noindent \textbf{Pre-trained Representations.} Pre-trained visual representations have become popular for promoting generalization in robot learning. To assess their impact on composition, we additionally evaluate using R3M \citep{nair2022r3m} and VC-1 \citep{majumdar2023we} as frozen visual encoders. We evaluate on a subset of factor pairs from our full pairwise evaluation. Specifically, we consider \emph{object type + table texture} (the pair of factors that affect physical motion the least), \emph{object position + table texture} (a pair where one factor is \emph{physical} and the other is \emph{visual}), and \emph{object position + table height} (a pair where both factors are \emph{physical}).

In our results in \cref{tab:r3m_results}, we find that both R3M and VC-1 perform poorly, with VC-1 failing to complete the task at all. Qualitatively, we observe that the trajectories are more jittery than when learning end-to-end, as also noted in prior work that tried R3M with diffusion-based policies \citep{chi2023diffusionpolicy}. This suggests that modern frozen visual representations are not effective for compositional generalization on real robots, and that end-to-end learning with prior robot data is superior for this.

\setlength{\tabcolsep}{4pt}
\begin{table}[t]
    \vspace{7pt}
    \begin{adjustbox}{width=\columnwidth}
    \begin{tabular}{l|cccc|ccc}
        \toprule
        Data Strategy                            & \multicolumn{4}{c|}{\textbf{L}}             & \multicolumn{3}{c}{\textbf{No Variation}} \\
        \midrule
        Train Method                             & Bridge         & R3M  & VC-1 & From Scratch & Bridge & R3M  & VC-1 \\
        \midrule
        \makecell[l]{Object Type +\\Table Tex}   & \textbf{8/9}   & 1/9  & 0/9  & 4/9          & 2/9    & 0/9  & 0/9  \\
        \makecell[l]{Object Pos +\\Table Tex}    & \textbf{5/9}   & 0/9  & 0/9  & 1/9          & 1/9    & 0/9  & 0/9  \\
        \makecell[l]{Object Pos +\\Table Height} & \textbf{2/9}   & 1/9  & 0/9  & 0/9          & 1/9    & 0/9  & 0/9  \\
        \midrule
        Overall                                  & \textbf{15/27} & 2/27 & 0/27 & 5/27         & 4/27   & 0/27 & 0/27 \\
        \bottomrule
    \end{tabular}
    \end{adjustbox}
    \vspace{-2pt}
    \caption{\small{Additional real robot pairwise composition results for our ``\emph{put fork in container}" task with R3M and VC-1.}}
    \label{tab:r3m_results}
    \vspace{-20pt}
\end{table}

\begin{figure*}[t]
\centering
\includegraphics[width=\textwidth]{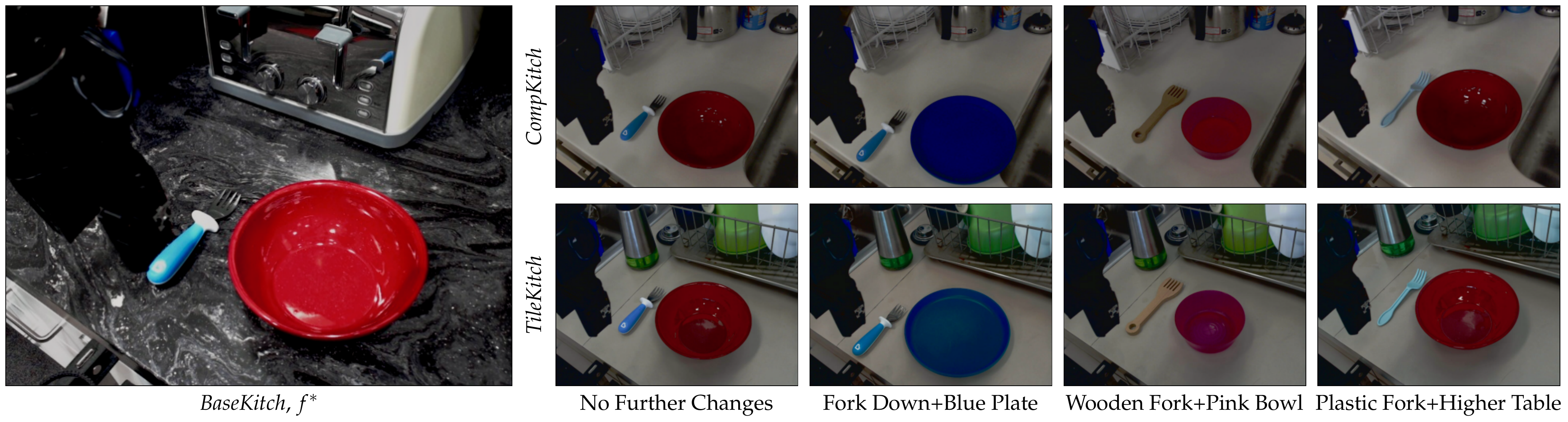}
\vspace{-15pt}
\caption{\small{Visualization of \emph{BaseKitch} with base factor values $f^*$ (left), compared to our transfer conditions in \emph{CompKitch} (top right) and \emph{TileKitch} (bottom right), which inherently have different factor values (e.g., \emph{table texture}) and other characteristics (e.g., \emph{distractor objects}, \emph{lighting}). We consider setups both with no further changes from $f^*$ beyond these inherent differences, as well as with additional changes.}}
\label{fig:transfer_kitchen}
\vspace{-5pt}
\end{figure*}

\begin{table*}[t]
    \begin{adjustbox}{width=\textwidth}
        \centering
        \begin{tabular}{l|l|ccc|ccc|ccc}
            \toprule
            Data Strategy & & \multicolumn{3}{c|}{\textbf{Stair}} & \multicolumn{3}{c|}{\textbf{L}} & \multicolumn{2}{c}{\textbf{No Variation}} \\
            \midrule
            Train Method  & & \makecell{Bridge\\(Co-FT)} & \makecell{Bridge\\(FT)} & \makecell{From\\Scratch} & \makecell{Bridge\\(Co-FT)} & \makecell{Bridge\\(FT)} & \makecell{From\\Scratch} & \makecell{Bridge\\(Co-FT)} & \makecell{Bridge\\(FT)} \\
            \midrule
            \multirow{2}{*}{\makecell[l]{No Further \\ Changes}}          & \emph{CompKitch} & 4/5            & 4/5          & 0/5  & \textbf{5/5} & 0/5          & 0/5          & 0/5   & 0/5  \\
                                                                          & \emph{TileKitch} & \textbf{5/5}   & 2/5          & 0/5  & \textbf{5/5} & \textbf{5/5} & 3/5          & 0/5   & 0/5  \\[2pt]
            \multirow{2}{*}{\makecell[l]{Fork Down + \\ Blue Plate}}      & \emph{CompKitch} & \textbf{4/5}   & \textbf{4/5} & 0/5  & 3/5          & 0/5          & 0/5          & 1/5   & 0/5  \\
                                                                          & \emph{TileKitch} & 4/5            & 3/5          & 0/5  & 3/5          & \textbf{5/5} & \textbf{5/5} & 0/5   & 0/5  \\[2pt]
            \multirow{2}{*}{\makecell[l]{Wooden Fork + \\ Pink Bowl}}     & \emph{CompKitch} & 3/5            & \textbf{5/5} & 0/5  & 2/5          & 0/5          & 0/5          & 0/5   & 0/5  \\
                                                                          & \emph{TileKitch} & 4/5            & 3/5          & 0/5  & 3/5          & \textbf{5/5} & 2/5          & 0/5   & 0/5  \\[2pt]
            \multirow{2}{*}{\makecell[l]{Plastic Fork + \\ Higher Table}} & \emph{CompKitch} & \textbf{4/5}   & 0/5          & 0/5  & 1/5          & 0/5          & 0/5          & 0/5   & 0/5  \\
                                                                          & \emph{TileKitch} & \textbf{3/5}   & 0/5          & 0/5  & 2/5          & 0/5          & 1/5          & 0/5   & 0/5  \\
            \midrule
            \multirow{3}{*}{Overall}                                      & \emph{CompKitch} & \textbf{15/20} & 13/20        & 0/20 & 11/20        & 0/20         & 0/20         & 1/20  & 0/20 \\
                                                                          & \emph{TileKitch} & \textbf{16/20} & 8/20         & 0/20 & 13/20        & 15/20        & 11/20        & 0/20  & 0/20 \\
                                                                          & Combined         & \textbf{31/40} & 21/40        & 0/40 & 24/40        & 15/40        & 11/40        & 1/40  & 0/40 \\
            \bottomrule
        \end{tabular}
    \end{adjustbox}
    \caption{\small{Out-of-domain transfer results to new kitchens \emph{CompKitch} and \emph{TileKitch}. We find that varied in-domain data from \emph{BaseKitch}, and BridgeData V2 as prior data, are both critical for effective transfer to these new kitchens. \textbf{Stair} outperforms \textbf{L}, although both achieve significant levels of transfer. Co-fine-tuning generally performs better than only fine-tuning.}}
    \label{tab:transfer_results}
    \vspace{-17.5pt}
\end{table*}

\vspace{-2pt}
\subsection{Out-Of-Domain Transfer}
\label{sec:ood_transfer}
To further demonstrate the utility of exploiting composition during data collection, we assess the transfer abilities of policies trained on our datasets to entirely new environments that capture some of the factor variety accounted for during data collection. We evaluate in two new kitchens, which we refer to as \emph{CompKitch} and \emph{TileKitch}.\footnote{We name these kitchens based on the material of their countertops. See \cref{fig:kitchen_views} in \cref{sec:real_eval_details} for additional views that make this more apparent.} These kitchens inherently have some completely out-of-distribution factor values from those studied in \emph{BaseKitch} (e.g., \emph{table texture}, shown in~\cref{fig:transfer_kitchen}), including for some factors we do not account for during data collection (e.g., \emph{distractor objects}, \emph{lighting}).

We first evaluate in these new kitchens with no further changes from the base factor values $f^*$ in \emph{BaseKitch}, aside from their inherent shifts. In addition, to assess beyond pairwise composition for multiple factors, we also evaluate in these kitchens with additional combinations of factor shifts. We visualize these different conditions in \cref{fig:transfer_kitchen}. For each data collection strategy, we again compare using BridgeData V2 as prior data, and training from scratch. In addition to using prior data for both pre-training and co-fine-tuning (as done with the previously evaluated models), we also compare to fine-tuning on only in-domain data from a pre-trained BridgeData V2 policy, to assess the importance of co-fine-tuning for compositional generalization and transfer.

We report our results in \cref{tab:transfer_results}. We find that BridgeData V2 is critical for transfer, as one of the two policies trained from scratch is unable to transfer at all, while the other only achieves a success rate of \textbf{11/40}. When co-fine-tuning with BridgeData V2, having varied in-domain data is still needed for robust transfer, as \textbf{No Variation} only achieves a success rate of \textbf{1/40}. These results indicate that this transfer setting represents a significant and challenging domain shift. 

Despite this, \textbf{Stair} and \textbf{L} both achieve significant levels of transfer, with success rates of \textbf{31/40} and \textbf{24/40}, respectively. \textbf{Stair} generally outperforms \textbf{L}, as was the case in simulation. This suggests that policies that use prior data are able to effectively transfer to new settings that require composition. When fine-tuning from a pre-trained model, \textbf{Stair} and \textbf{L} produce policies that achieve some transfer, but less consistently than when co-fine-tuning, with reduced success rates of \textbf{21/40} and \textbf{15/40}, respectively. In particular, the \textbf{L} policy fails to transfer to \emph{CompKitch}, and both policies fail to transfer to \textbf{Plastic Fork + Higher Table} in both kitchens. This suggests that co-fine-tuning is generally superior for facilitating composition than pre-training alone. Overall, we believe these results further suggest that policies can exhibit composition to generalize to unseen settings, our data collection strategies are sufficient to achieve some of this composition, and that prior data is important for this composition to happen effectively.

\subsection{Additional Experiments}
\begin{figure}[t]
    \centering
    \begin{subfigure}[t]{0.32\columnwidth}
        \includegraphics[width=\textwidth]{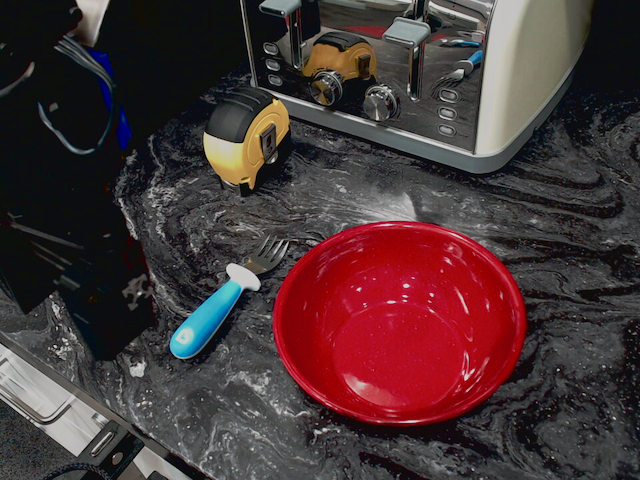}
        \caption{Tape Measure}
    \end{subfigure}
    \begin{subfigure}[t]{0.32\columnwidth}
        \includegraphics[width=\textwidth]{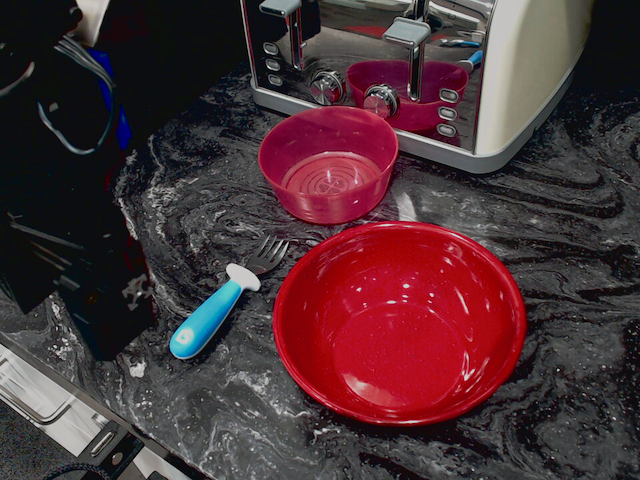}
        \caption{Pink Bowl}
    \end{subfigure}
    \begin{subfigure}[t]{0.32\columnwidth}
        \includegraphics[width=\textwidth]{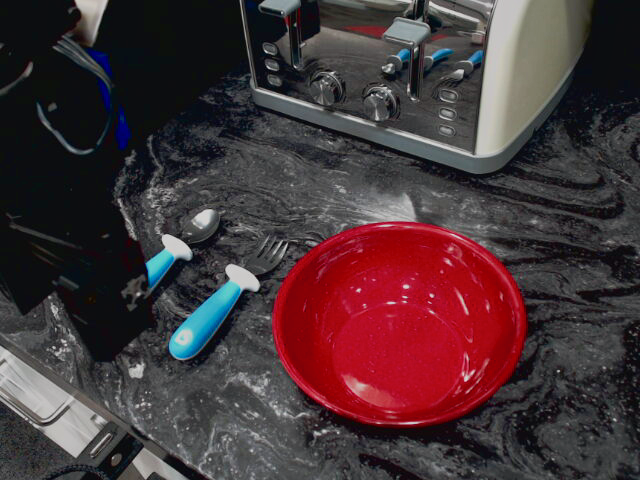}
        \caption{Spoon}
    \end{subfigure}
    \caption{\small{Visualization of the different values for \emph{distractor objects} we consider in our held-out factor evaluation.}}
    \label{fig:distractors}
    \vspace{-20pt}
\end{figure}

\noindent \textbf{Unaccounted Factors.}
Our transfer experiments involve generalization to some factors that were unaccounted for during data collection, such as \emph{distractor objects} and \emph{lighting}. For more focused assessment of the impact of data collection strategies on such unaccounted factors, we additionally evaluate on \emph{distractor objects} as a held-out factor in \emph{BaseKitch}. In early experiments, we noticed a large degree of robustness to this factor from \textbf{No Variation} policies, so we did not include it among our primary factors. We evaluate our policies on 3 different values for this factor (consisting of distinct objects and positions for each value), which we visualize in \cref{fig:distractors}. Our results in \cref{table:distractor_eval} confirm the aforementioned robustness of the \textbf{No Variation} policies. However, when training from scratch, policies using \textbf{Stair} and \textbf{L} data perform worse, indicating that these data collection strategies may worsen robustness for unaccounted factors, possibly due to introducing spurious correlations. However, incorporating prior data completely mitigates this issue and restores robustness to this factor. This suggests another reason for why using prior robot data can be important for generalization: to address factors unaccounted for during in-domain data collection.

\begin{table}[t]
    \centering
    \vspace{5pt}
    \begin{adjustbox}{width=\columnwidth}
    \begin{tabular}{l|cc|cc|cc}
        \toprule
        Data Strategy & \multicolumn{2}{c|}{\textbf{Stair}}        & \multicolumn{2}{c|}{\textbf{L}}             & \multicolumn{2}{c}{\textbf{No Variation}}  \\
        \midrule
        Train Method  & Bridge          & \makecell{From\\Scratch} & Bridge         & \makecell{From\\Scratch}   & Bridge          & \makecell{From\\Scratch} \\
        \midrule
        Tape Measure  & 5/5             & 1/5                      & 5/5            & 5/5                        & 5/5             & 5/5                      \\
        Pink Bowl     & 5/5             & 3/5                      & 5/5            & 4/5                        & 5/5             & 5/5                      \\
        Spoon         & 5/5             & 5/5                      & 5/5            & 0/5                        & 5/5             & 4/5                      \\
        \midrule
        Overall       & \textbf{15/15}  & 9/15                     & \textbf{15/15} & 9/15                       & \textbf{15/15}  & \textbf{15/15}           \\
        \bottomrule
    \end{tabular}
    \end{adjustbox}
    \caption{\small{Evaluation on held-out factor \emph{distractor objects}. Data collection strategies that do not account for this factor can perform worse than \textbf{No Variation} when training from scratch, but this issue is alleviated completely by using prior data.}}
    \label{table:distractor_eval}
    \vspace{-5pt}
\end{table}

\begin{figure}[t]
    \vspace{5pt}
    \centering
    \begin{subfigure}[t]{0.48\columnwidth}
        \includegraphics[width=\textwidth]{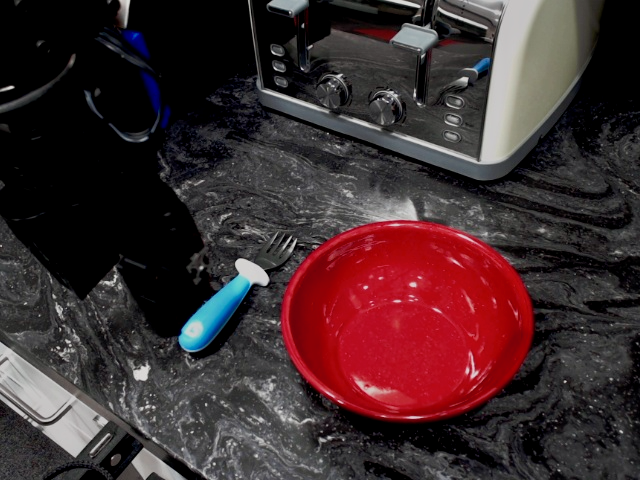}
        \caption{Main Camera}
    \end{subfigure}
    \begin{subfigure}[t]{0.48\columnwidth}
        \includegraphics[width=\textwidth]{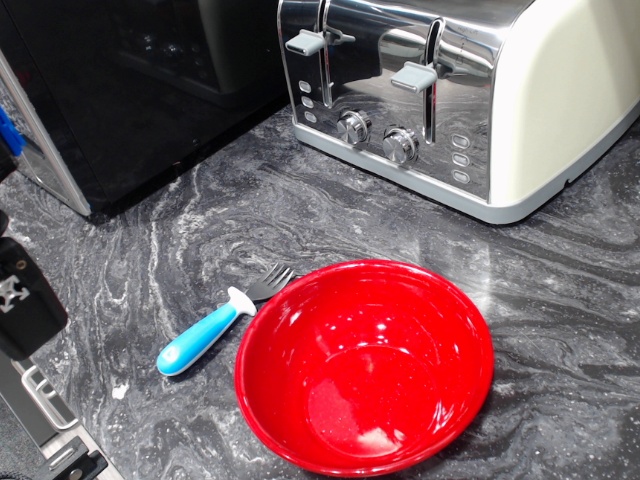}
        \caption{Secondary Camera}
    \end{subfigure}
    \caption{\small{Visualization of the secondary camera view we use for studying composition of \emph{camera position} (right), compared with our main camera view used in all other experiments (left).}
    \label{fig:second_cam}}
    \vspace{-17.5pt}
\end{figure}

\smallskip \noindent \textbf{Camera Position Composition.}
We do not consider \emph{camera position} among the primary factors in our real evaluation, due to the difficulty of varying this factor in a controlled manner. However, we have one additional third-person camera on our robot platform that we also collect data for, which allows us to separately study compositional generalization with this secondary camera. We visualize the difference in perspective with this secondary camera from our main camera in \cref{fig:second_cam}.

We focus on the composition of \emph{camera position} with \emph{table texture}. We evaluate in \emph{BaseKitch}, but using the secondary camera for observations. We set \emph{table texture} to the value \textbf{white marble} (third row in \cref{fig:real_factors}), as this was the value that was the most challenging for policies lacking variation for \emph{table texture} to generalize to. We train a policy on the combination of two sub-datasets: our \textbf{L} dataset from the primary camera view, and the same \textbf{L} dataset but from the secondary camera view, with the exception of data for \emph{table texture}. We compare against training on only one of these sub-datasets, which either lack data with the value for \emph{camera position} or \emph{table texture} that is seen during evaluation. We again evaluate how BridgeData V2 affects this composition.

In our results in \cref{tab:cam_results}, we find that yet again, policies that do not use prior data struggle to generalize. When using prior data, training on each sub-dataset individually results in a policy that sometimes generalizes, but not consistently. However, training on both sub-datasets together is able to achieve a perfect success rate, suggesting effective composition of \emph{camera position} with \emph{table texture}.

\begin{table}[h]
    \centering
    \vspace{-5pt}
    \begin{tabular}{l|cc}
        \toprule
        Train Method       & Bridge         & From Scratch \\
        \midrule
        No Camera Position & 1/10           & 0/10         \\
        No Table Texture   & 6/10           & 0/10         \\
        Both               & \textbf{10/10} & 0/10         \\
        \bottomrule
    \end{tabular}
\caption{\small{Our policy composes datasets missing either the correct \emph{camera position} or \emph{table texture} seen during evaluation, to outperform training on either dataset alone. Prior data is critical for composition.}}
\label{tab:cam_results}
\vspace{-10pt}
\end{table}

\vspace{-3pt}
\section{Discussion}
\label{sec:discussion}

\subsection{Summary}
We investigate the compositional abilities of visual imitation learning policies for robotic manipulation, and whether this can be exploited by data collection strategies to more efficiently achieve broad generalization. In summary, our simulated and real-world experiments suggest the following:
\begin{itemize}
    \item Robot policies do exhibit significant composition, although the degree of composition varies across different factor pairs, with composition between \emph{physical} factors generally being the most challenging.
    \item Leveraging prior robot data is critical for strengthening compositional abilities with a real robot, and co-fine-tuning is generally better for this than fine-tuning.
    \item Our proposed data collection strategies are able to exploit composition to reduce data collection effort, while producing policies that can generalize to unseen settings.
    \item Two of our proposed strategies, \textbf{Stair} and \textbf{L}, are able to transfer policies to entirely new environments that require a high degree of compositional generalization, showing they can produce data that is useful beyond the original domain it was collected in.
    \item Our proposed \textbf{Stair} strategy is generally the most effective, as it achieves the best pairwise composition in simulation and real, the best transfer results in real, and is the only strategy in simulation that effectively scales with dataset size for the same amount of factor variation.

\end{itemize}
Overall, we believe these results provide insights on how roboticists can more efficiently collect in-domain data for achieving generalization in their desired settings.

\subsection{Limitations and Future Work}
\smallskip \noindent \textbf{More Robot Platforms and Tasks.}
While we conduct thorough real-world experiments for evaluating compositional generalization, it would be interesting to scale our experiments and analysis to more complex robot platforms and tasks. We did some preliminary investigation into treating different tasks as a factor for composition, by training a policy on the combination of the \textbf{L} dataset for our main task ``\emph{put fork in container}", and \textbf{No Variation} data for a new task ``\emph{remove fork from container}". We then evaluated this on the new task with factor values found in the main task data, but it did not succeed. This could be because our new task data did not have any factor variation, which might be necessary for composition. Future work can more thoroughly consider this setting.

\smallskip \noindent \textbf{Large-Scale Data Collection.} Although we believe our results can be broadly informative for data collection in robotics, we focus primarily on in-domain data. Future work could scale this analysis to large-scale data collection efforts, to better understand how to most efficiently collect prior data for downstream transfer. While we do show that incorporating such prior data is critical for strengthening composition, we only consider one prior dataset. Future work can investigate how other prior robotic datasets impact composition, and what aspects of prior datasets are the most important for this.

\smallskip \noindent \textbf{Improving Composition.} While our results show that robot policies can possess significant composition and transfer capabilities, they still struggle at times with composition, particularly for factors that interact physically. Future work can better address generalization in these settings, such as by prioritizing data collection for certain factor combinations over others. Also, we only consider straightforward behavior cloning methods for policy learning. It would be interesting to study if other policy learning methods have better compositional abilities, such as different learning algorithms or architectures.

For example, one possible direction for future work could involve conditioning diffusion-based policies on individual environmental factors. Then, the score predictions for separate factors could be combined to produce a policy that more effectively achieves composition, similar to prior work in the context of text-to-image generation \citep{liu2022compositional}. Prior work has studied combining score predictions in this manner via classifier-free guidance \citep{ho2022classifier} in the context of goal-conditioned robotic policy learning \citep{reuss2023goal}, but this direction has not yet been studied for composing environmental factors.

\section*{Acknowledgments}
This work was supported by DARPA W911NF2210214, ONR N00014-22-1-2293, NSF 1941722, Toyota Research Institute, and Other Transaction award HR00112490375 from the U.S. Defense Advanced Research Projects Agency (DARPA) Friction for Accountability in Conversational Transactions (FACT) program. We thank Suvir Mirchandani, Joey Hejna, and other members of the ILIAD lab for helpful discussions and feedback. We also thank Abraham Lee for advice on setting up our robot platform to work with BridgeData V2.

\bibliographystyle{plainnat}
\bibliography{references}

\clearpage
\begin{figure*}[t]
    \centering
    \includegraphics[width=\textwidth]{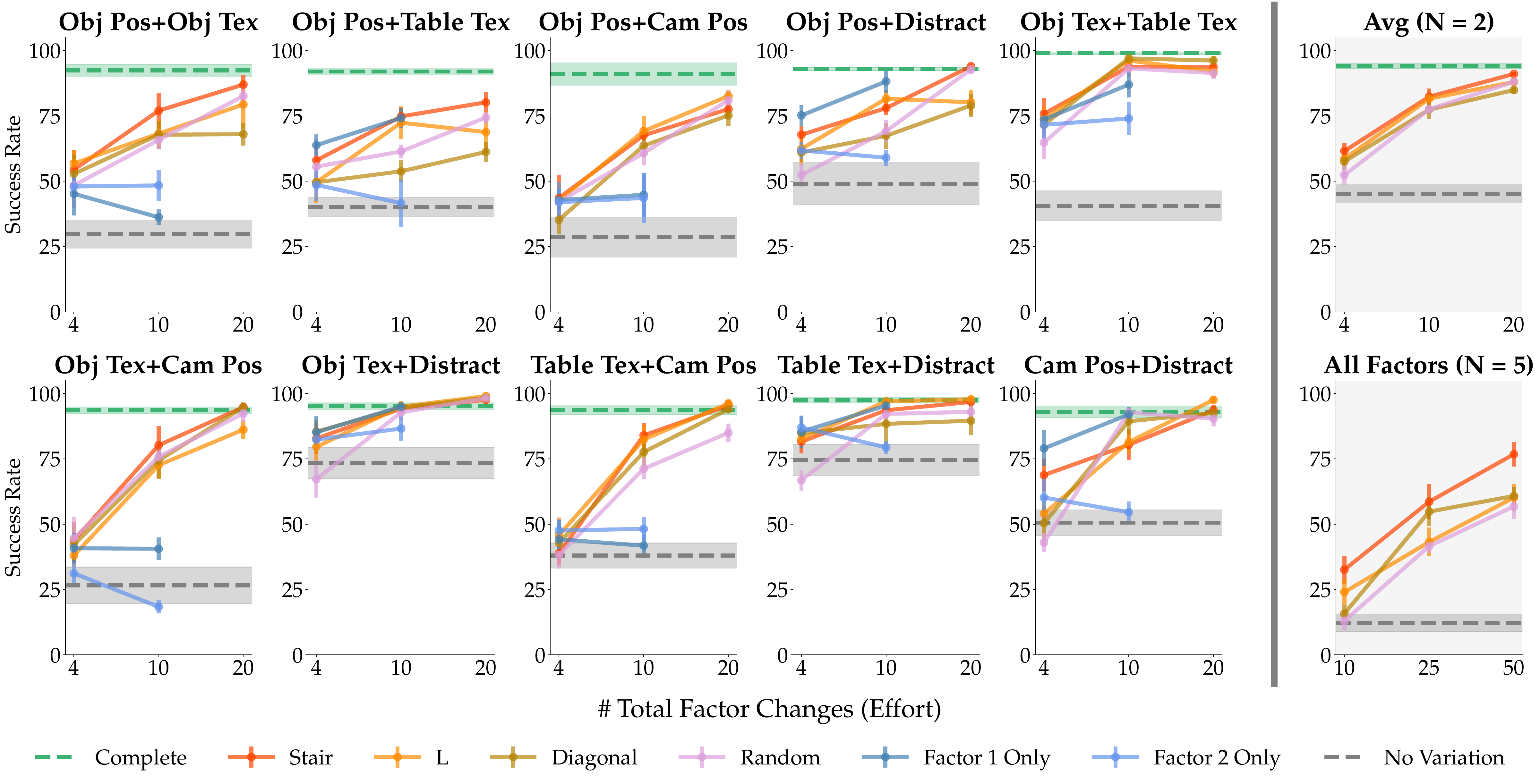}
    \caption{
    \small{Simulation results of different data collection strategies for \emph{Door Open}. We report results where $\mathcal{F}^N$ consists of each possible factor pair ($N = 2$), average results across all pairs, and results where $\mathcal{F}^N$ consists of all factors ($N = 5$). Results are similar as with \emph{Pick Place}, with \textbf{Stair} generally performing the best, especially in the ($N = 5$) setting. Error bars represent standard error across 5 seeds.}}
    \label{fig:door_open_results}
    \vspace{-10pt}
\end{figure*}

\appendices

\section{Data Collection Strategies}
\subsection{Pseudocode}
\label[appendix]{sec:strat_pseudocode}
We provide pseudocode for the data collection strategies proposed in \cref{sec:data_strats} that are intended to exploit compositional generalization (\textbf{Diagonal}, \textbf{L}, \textbf{Stair}).

\floatname{algorithm}{Algorithm}
\vspace{-5pt}
\begin{algorithm}
\caption{\textbf{Diagonal}}\label{alg:complete}
\begin{algorithmic}
\State \textbf{Input:} scene \textbf{S}, factor values \textbf{F} (size $N$ factors $\times$ $k$ values)

\For{$j \gets 1$ to $k$}
    \State $f \gets 0^N$
    \For{$i \gets 1$ to $N$}
        \State $f_{i} \gets \textbf{F}_{ij}$
    \EndFor
    \State \Call{SetFactors}{\textbf{S}, $f$}
    \State \Call{CollectData}{\textbf{S}}
\EndFor
\end{algorithmic}
\end{algorithm}

\vspace{-15pt}
\begin{algorithm}
\caption{\textbf{L}}\label{alg:l}
\begin{algorithmic}
\State \textbf{Input:} scene \textbf{S}, factor values \textbf{F} (size $N$ factors $\times$ $k$ values), base factor values $f^*$ (size $N$ factors)

\For{$i \gets 1$ to $N$}
    \State $f \gets f^*$
    \For{$j \gets 1$ to $k$}
        \State $f_{i} \gets \textbf{F}_{ij}$
        \State \Call{SetFactors}{\textbf{S}, $f$}
        \State \Call{CollectData}{\textbf{S}}
    \EndFor
\EndFor
\end{algorithmic}
\end{algorithm}

\begin{algorithm}
\caption{\textbf{Stair}}\label{alg:stair}
\begin{algorithmic}
\State \textbf{Input:} scene \textbf{S}, factor values \textbf{F} (size $N$ factors $\times$ $k$ values), base factor values $f^*$ (size $N$ factors)
\State $f \gets f^*$
\For{$j \gets 1$ to $k$}
    \For{$i \gets 1$ to $N$}
        \State $f_{i} \gets \textbf{F}_{ij}$
        \State \Call{SetFactors}{\textbf{S}, $f$}
        \State \Call{CollectData}{\textbf{S}}
    \EndFor
\EndFor
\end{algorithmic}
\end{algorithm}

\vspace{-10pt}
\section{Simulation Experiments}
\label[appendix]{sec:sim_eval_details}

\subsection{Door Open Results}
\label[appendix]{sec:door_open_results}

We include additional simulation results for the task \emph{Door Open} in \cref{fig:door_open_results}.
Results are similar as in \emph{Pick Place}, with generally strong pairwise composition, and \textbf{Stair} generally performing the best, especially in the $N = 5$ setting.

\vspace{-3pt}
\subsection{Factors}
\begin{figure*}[t]
    \centering
    \begin{subfigure}[t]{\textwidth}
        \centering
        \includegraphics[width=0.23\textwidth]{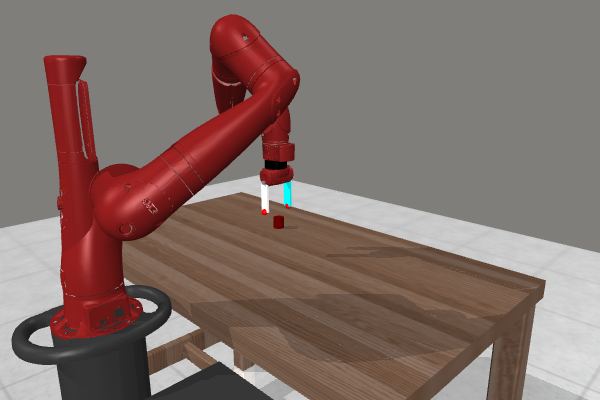}\hspace{1pt}
        \includegraphics[width=0.23\textwidth]{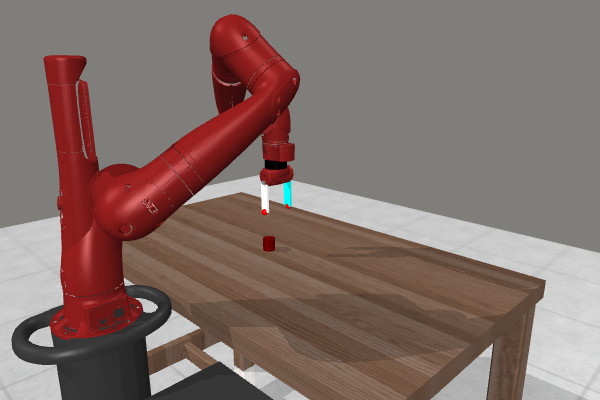}\hspace{10pt}
        \includegraphics[width=0.23\textwidth]{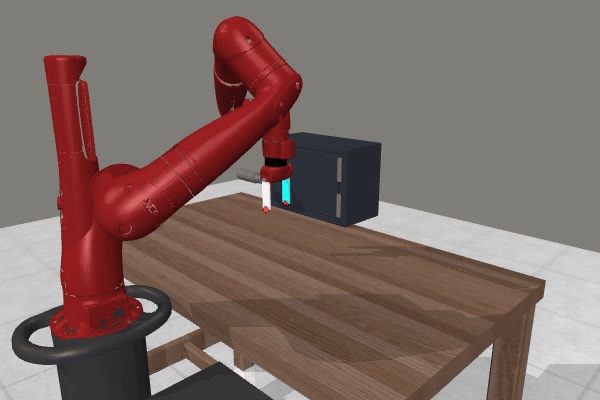}\hspace{1pt}
        \includegraphics[width=0.23\textwidth]{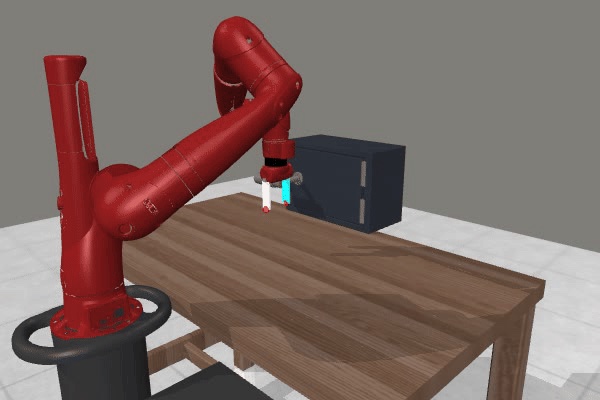}
        \caption{\emph{Object Position}}
        \vspace{5pt}
    \end{subfigure}
    \begin{subfigure}[t]{\textwidth}
        \centering
        \includegraphics[width=0.23\textwidth]{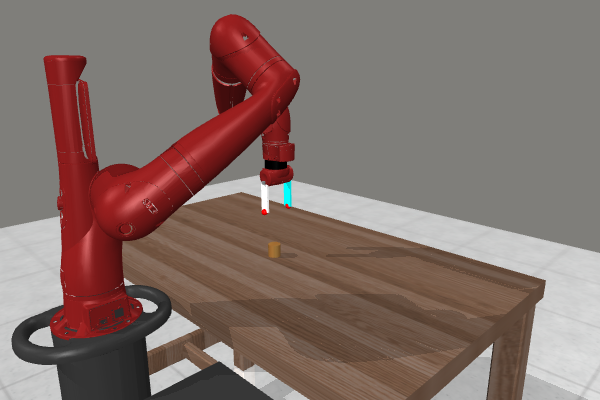}\hspace{1pt}
        \includegraphics[width=0.23\textwidth]{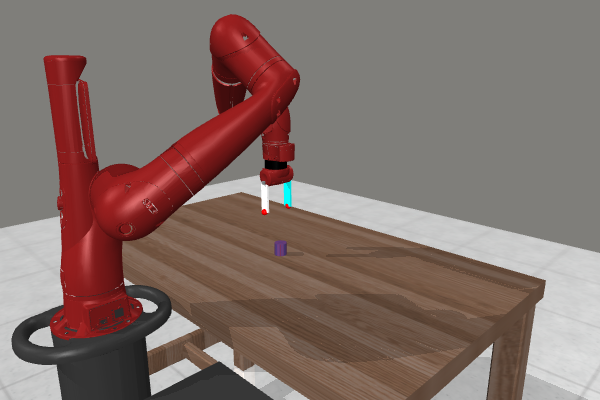}\hspace{10pt}
        \includegraphics[width=0.23\textwidth]{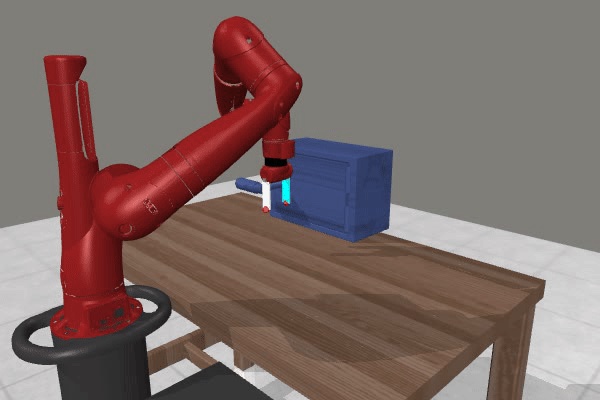}\hspace{1pt}
        \includegraphics[width=0.23\textwidth]{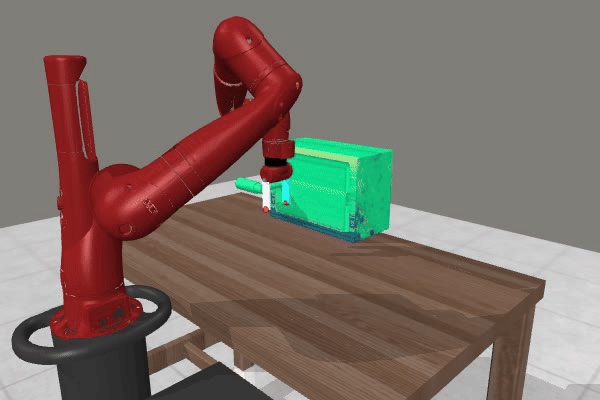}
        \caption{\emph{Object Texture}}
        \vspace{5pt}
    \end{subfigure}
    \begin{subfigure}[t]{\textwidth}
        \centering
        \includegraphics[width=0.23\textwidth]{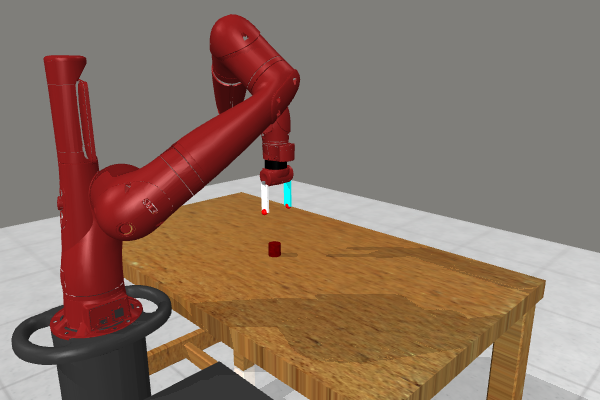}\hspace{1pt}
        \includegraphics[width=0.23\textwidth]{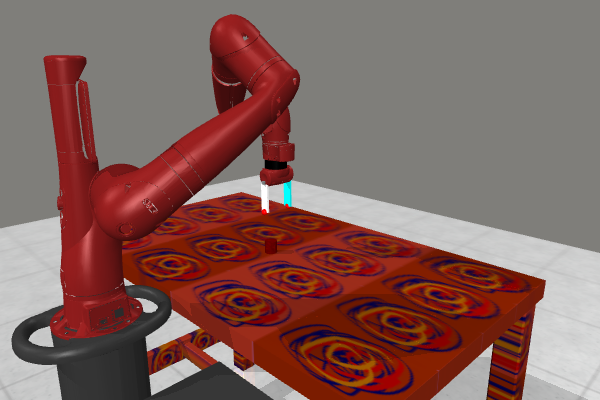}\hspace{10pt}
        \includegraphics[width=0.23\textwidth]{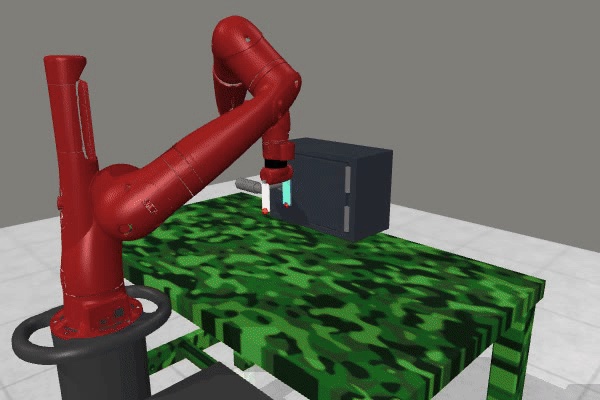}\hspace{1pt}
        \includegraphics[width=0.23\textwidth]{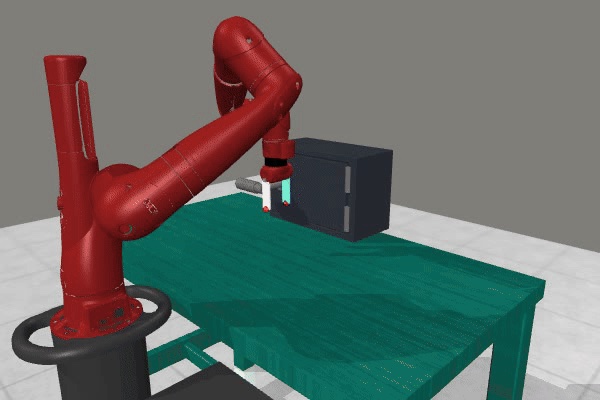}
        \caption{\emph{Table Texture}}
        \vspace{5pt}
    \end{subfigure}
    \begin{subfigure}[t]{\textwidth}
        \centering
        \includegraphics[width=0.23\textwidth]{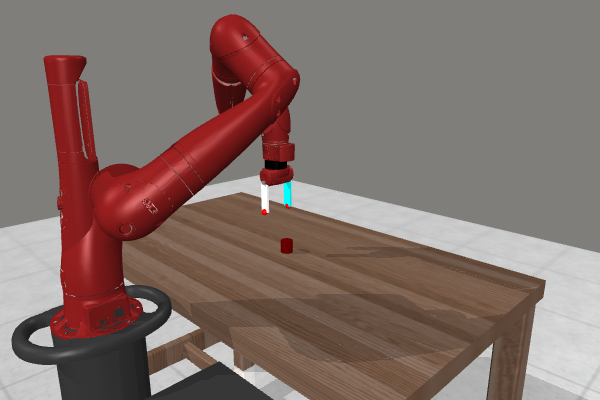}\hspace{1pt}
        \includegraphics[width=0.23\textwidth]{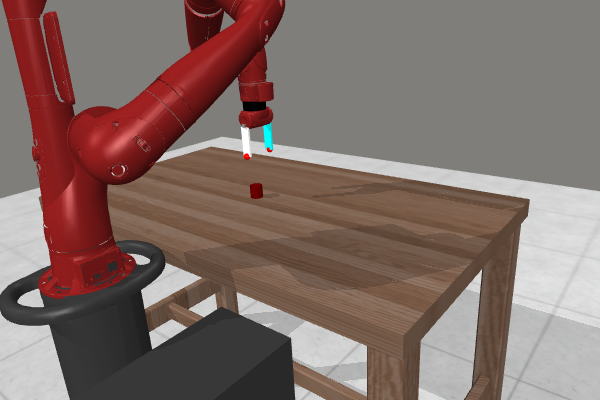}\hspace{10pt}
        \includegraphics[width=0.23\textwidth]{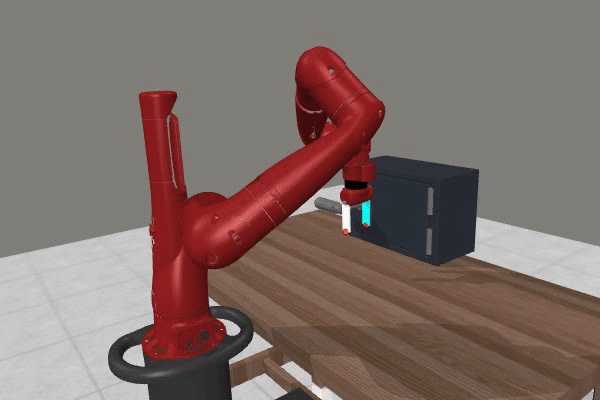}\hspace{1pt}
        \includegraphics[width=0.23\textwidth]{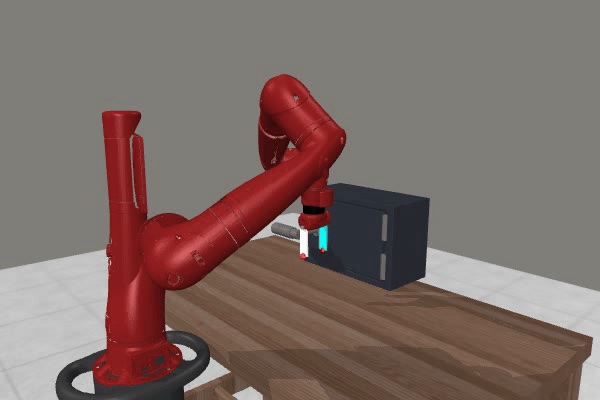}
        \caption{\emph{Camera Position}}
        \vspace{5pt}
    \end{subfigure}
    \begin{subfigure}[t]{\textwidth}
        \centering
        \includegraphics[width=0.23\textwidth]{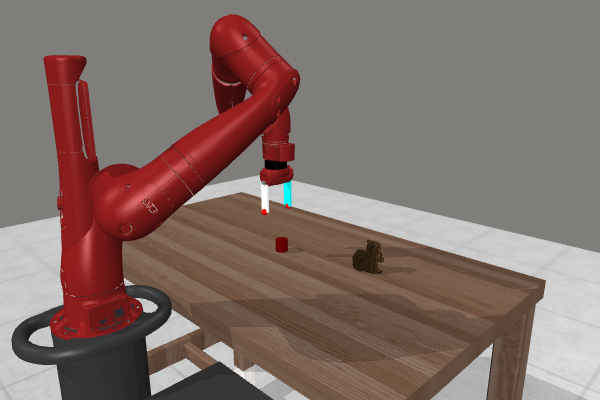}\hspace{1pt}
        \includegraphics[width=0.23\textwidth]{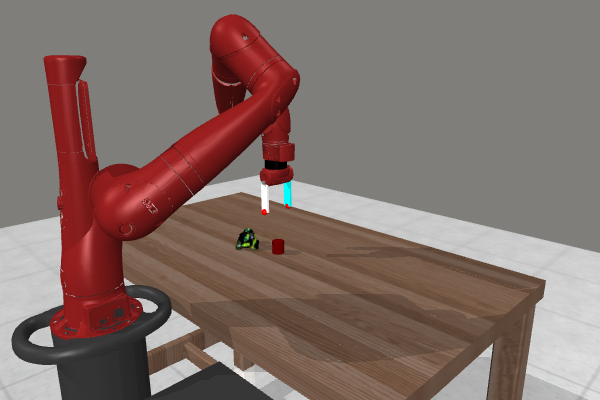}\hspace{10pt}
        \includegraphics[width=0.23\textwidth]{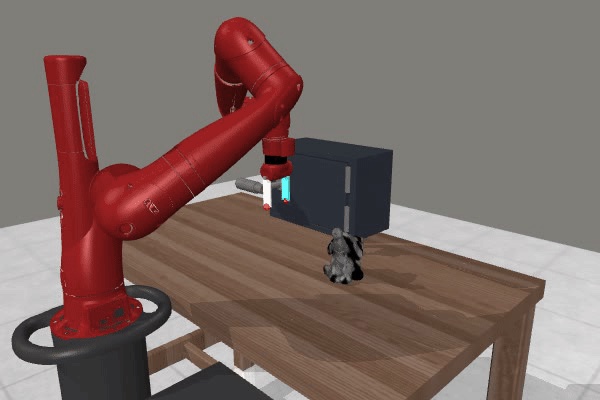}\hspace{1pt}
        \includegraphics[width=0.23\textwidth]{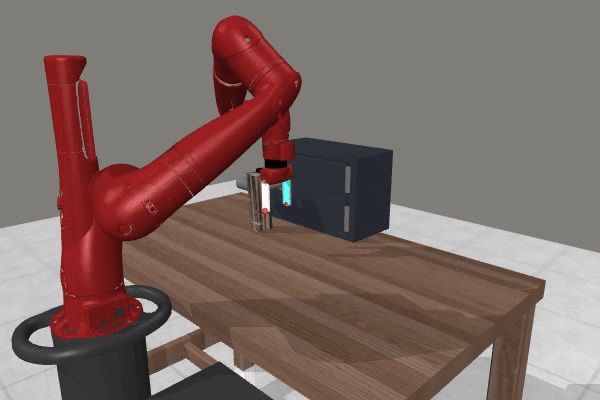}
        \caption{\emph{Distractor Objects}}
    \end{subfigure}
    \caption{\small{Visualization of our factors for the \emph{Pick Place} (left) and \emph{Door Open} (right) tasks from \emph{Factor World}. We show two examples of values for each factor we consider.}}
    \label{fig:sim_factors}
    \vspace{-7.5pt}
\end{figure*}

For the inherently discrete-valued factors \emph{object texture}, \emph{table texture}, and \emph{distractor objects}, we sample our $k=10$ values for $\mathcal{F}^N$ from all possible training values specified in \emph{Factor World}. \emph{Distractor objects} also include a size scale (sampled from range $[0.3, 0.8]$), 2D rotation (sampled from range $[0, 2\pi]$), and 2D position (sampled from all possible positions on the table) as part of each value. For \emph{object position}, we sample 2D $xy$ positions from the range $[-0.1, 0.1]$ for both coordinates. We note that the \emph{Pick Place} task includes a small amount of added noise to \emph{object position} each episode, sampled uniformly from the range $[-0.03, 0.03]$. For \emph{camera position}, we sample 3D $xyz$ positions and 4D rotation quaternions all from the range $[-0.05, 0.05]$. When sampling our $k$ values for each factor, we ensure that the scripted policy is able to solve the task for each value, because some values for \emph{object position} and \emph{distractor objects} can impede the task. In \cref{fig:sim_factors}, we visualize two examples of values for each factor, for both the \emph{Pick Place} and \emph{Door Open} tasks. We note that each random seed in our evaluation includes a different set of $k=10$ values sampled for each factor for $\mathcal{F}^N$.

\subsection{Training}
We use the same policy architecture and training hyperparameters from the original \emph{Factor World} experiments \citep{xie2023decomposing}. We condition policies on the same observations (2 84x84 RGB images from 2 camera views without history, and proprioception), and use the same action space (3D end-effector position deltas and open/close gripper). Unlike the original \emph{Factor World} experiments, we always use random shift augmentation. Unlike our real robot experiments, we train policies without goal conditioning, as we do not leverage diverse prior data in this setting, so task conditioning is not as essential.

\subsection{Color Jitter Augmentation}
\label[appendix]{sec:color_jitter}
\begin{figure*}[t]
    \centering
    \includegraphics[width=\textwidth]{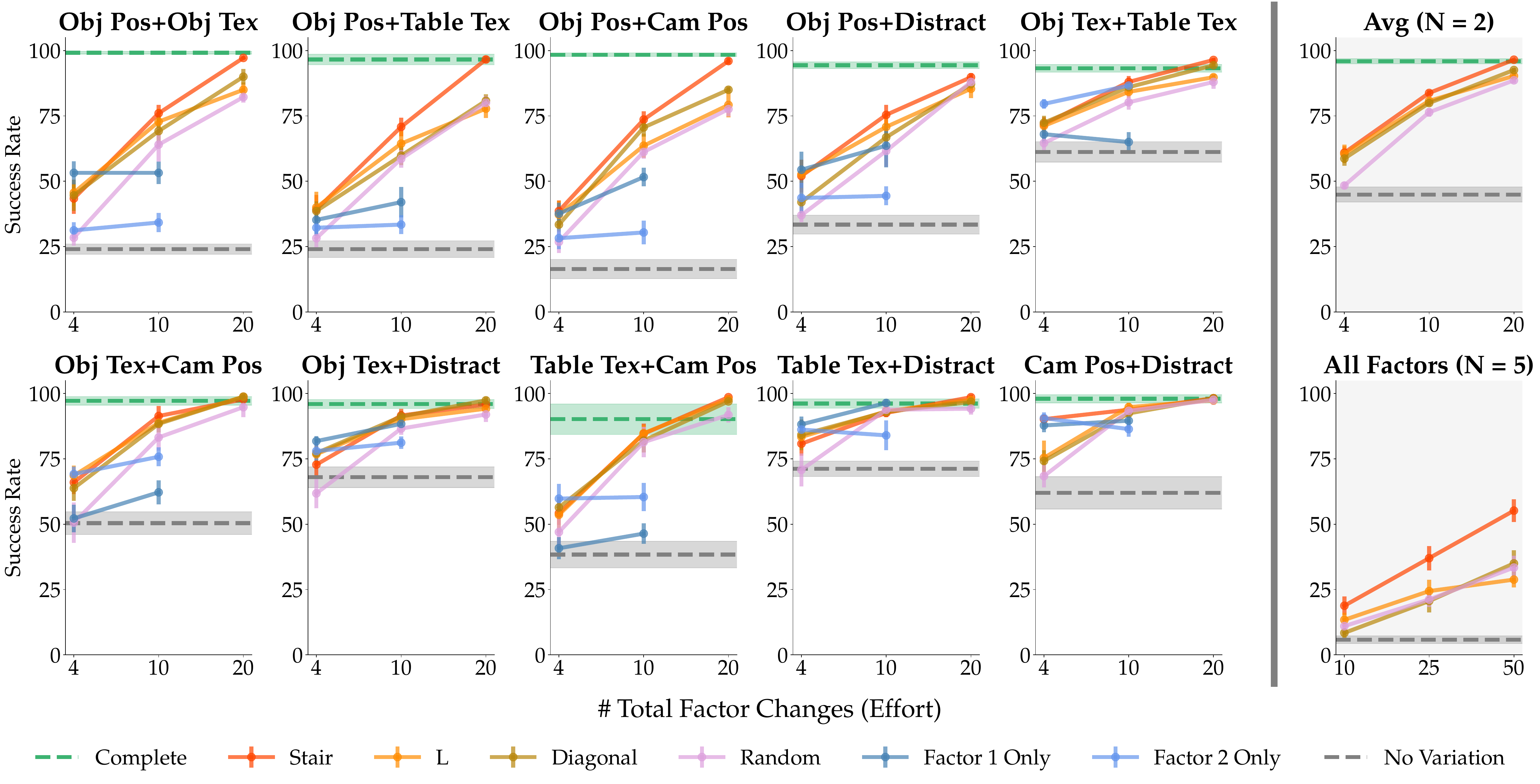}
    \caption{
    \small{Simulation results of data collection strategies for \emph{Pick Place}, with additional color jitter augmentation. Overall performance across all strategies slightly improves, but the overall trends are similar to as without color jitter. Error bars represent standard error across 5 seeds.}}
    \label{fig:pick_place_color}
    \vspace{-17.5pt}
\end{figure*}

We do not use color jitter augmentation in our main simulation experiments, because the original experiments for \emph{Factor World} found this to reduce overall generalization, except for when training variation was very low \citep{xie2023decomposing}. However, here we include additional results with color jitter (in addition to random shift). We find that overall performance across all strategies does slightly improve, but the overall trends across strategies are similar to as without color jitter.

\vspace{-5pt}
\subsection{Accounting for Factor Value Similarity}
\label[appendix]{sec:accounting_for_similarity}
\begin{figure}[t]
    \centering
    \vspace{5pt}
    \includegraphics[width=0.75\columnwidth]{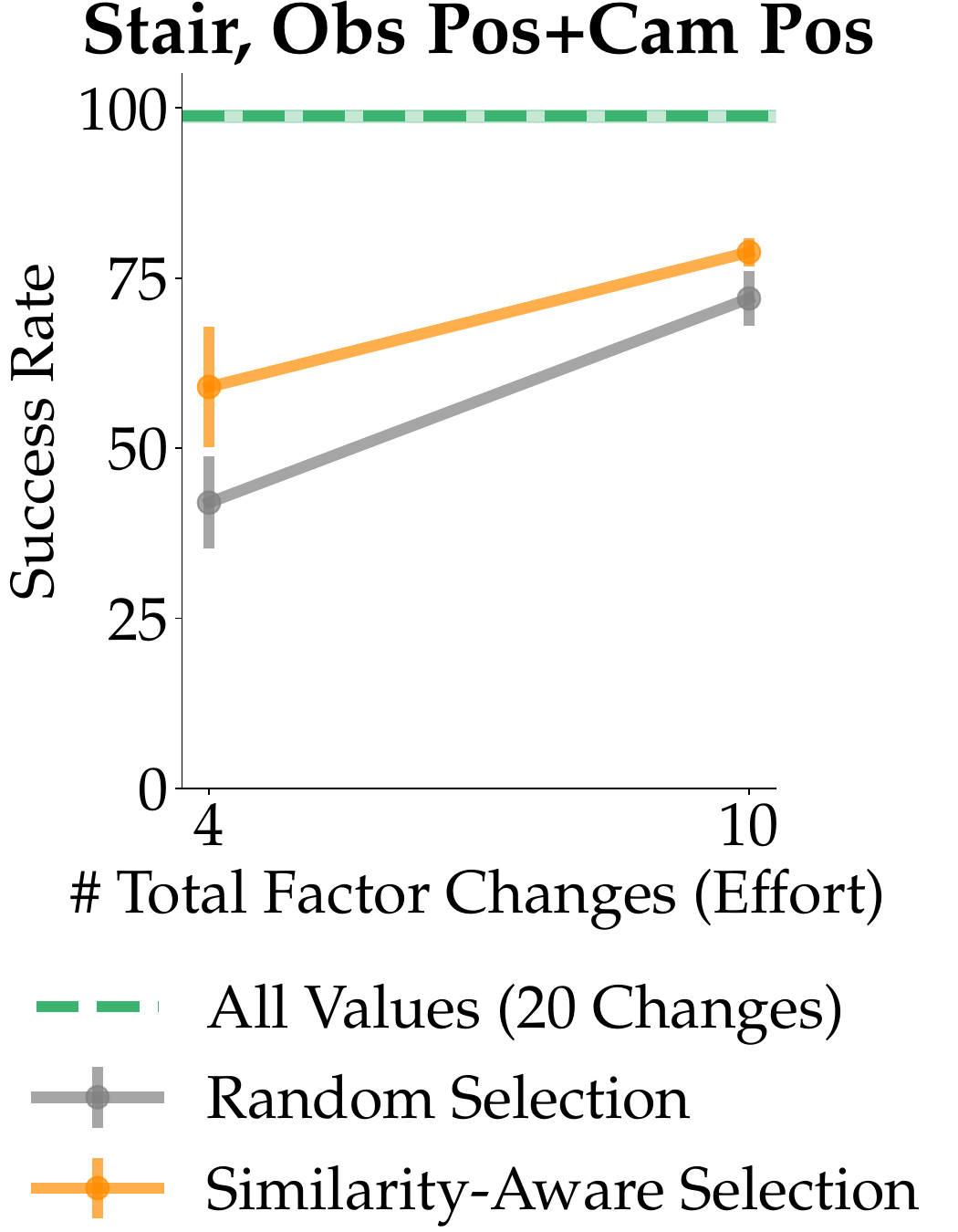}
    \vspace{-2pt}
    \caption{\small{Performance of the \textbf{Stair} strategy for the \emph{Pick Place} task, when composing \emph{object position} and \emph{camera position}. We find that selecting factor values using a similarity-aware strategy outperforms random selection (as done in our main simulation results).}}
    \label{fig:similarity_aware_data}
    \vspace{-19pt}
\end{figure}

In practice, data collection strategies should ideally account for similarity between factor values to improve generalization. For example, when training policies to be robust to the factor \emph{object type}, it would be desirable to prioritize object diversity during data collection to generalize to as many objects as possible, rather than collecting data for overly similar objects. Here, we demonstrate how our proposed data collection framework could incorporate notions of factor similarity when available, through additional \emph{Pick Place} experiments.

We consider the composition of \emph{object position} and \emph{camera position} in the $N = 2$ setting, as these are the only factors we study in simulation where computing similarity/distance is straightforward (we use Euclidean distance for 3D positions, and angular distance for camera rotation quaternions). We modify \textbf{Stair} to choose factor values using a similarity-aware strategy, rather than randomly as before. We do this by running a $k$-medoids algorithm on the set of 10 values for each factor, to determine which $k$ values/medoids in the set minimize the sum of distances from each value to its nearest medoid. We determine $k$ according to what is permissible under different factor change budgets (e.g., for 10 total changes in the $N = 2$ setting, each factor can have $k = \frac{10}{2} = 5$ values).

In \cref{fig:similarity_aware_data}, we compare using $k$-medoids for factor value selection (orange), and random selection as done in our main results (gray). We evaluate using the same procedure as in our main simulation results in \cref{sec:sim_results}, where the aim is to generalize to all factor value combinations. We find that accounting for similarity does indeed improve upon random selection, although it does not match observing all factor values (green), which requires 20 factor changes.

We note that it often not straightforward to compute similarity metrics for other factors, and that while accounting for similarity can be easily incorporated into our data collection framework, it is mostly orthogonal to our study of composition. We believe it would be interesting for future work to investigate methods of computing similarity metrics for other factors, such as by using text and/or image embeddings, and leveraging this to further improve data collection.

\vspace{-3pt}
\subsection{Computing Factor Changes}
For our results in \cref{sec:sim_results}, we compute the total number of factor changes for each strategy as follows. We assume the initial configuration of factor values requires $n$ changes, one for each factor. For \textbf{Stair}, \textbf{L}, and \textbf{Single Factor}, we assume each new configuration of factor values requires 1 additional change. We note this could be a slight underestimate for \textbf{L} in practice, as this strategy could require some additional changes when finishing varying one factor and returning to the base factor values $f^*$, to begin data collection for varying another factor. For \textbf{Diagonal} and \textbf{Random}, we assume each new configuration of factor values requires $n$ additional changes, as new values for all factors are resampled. We note that this could be a slight overestimate for \textbf{Random} in practice, as some of the resampled factor values may not change.

For each budget of factor value changes, we determine the amount of factor value configurations allowed for each strategy. We then divide the budget of total demonstrations by this to determine how many to collect for each configuration, with any remainder going to the last configuration.

\section{Real Robot Experiments}
\begin{figure*}[t]
    \centering
    \begin{subfigure}[t]{0.32\textwidth}
        \includegraphics[width=\textwidth]{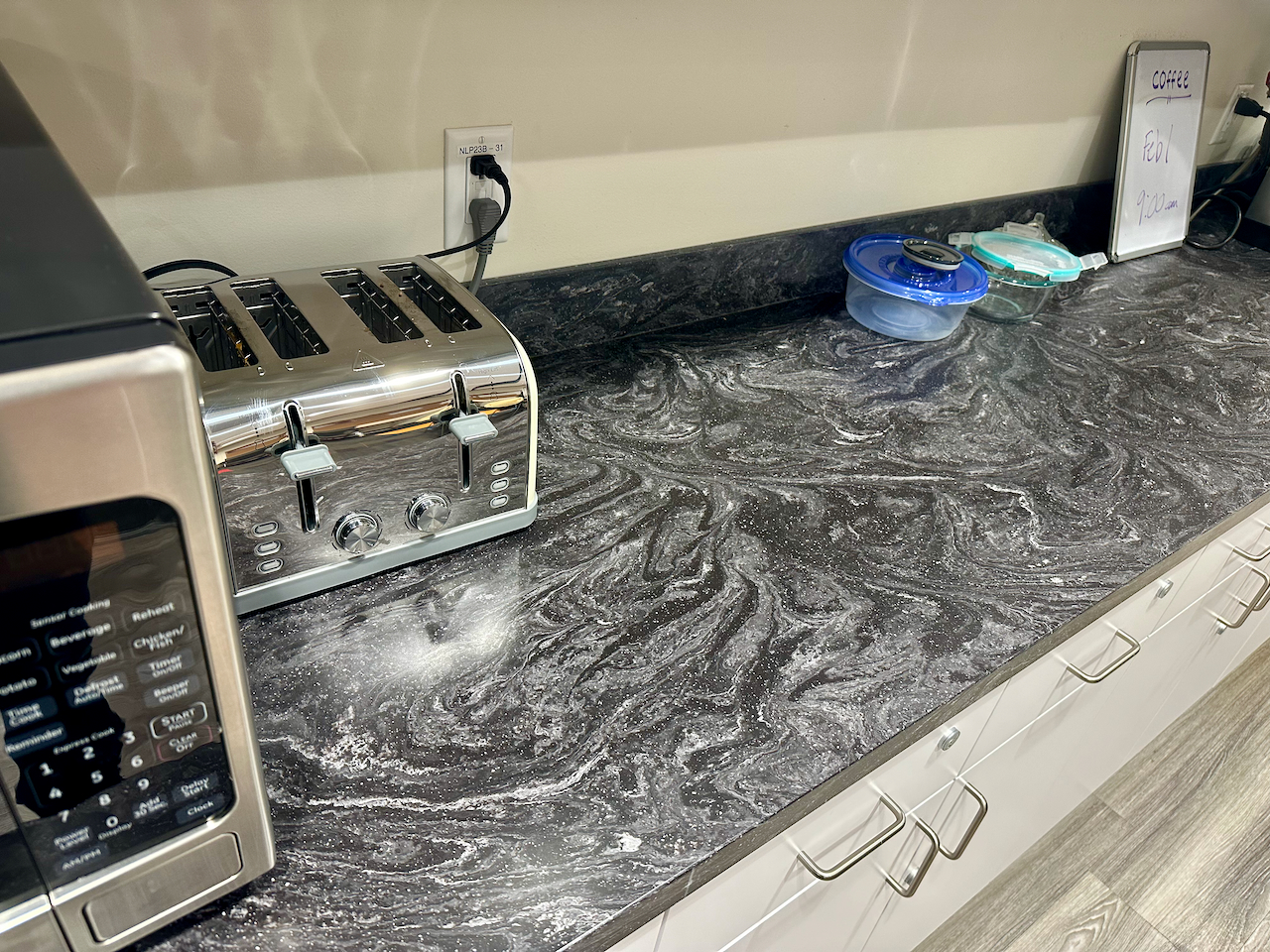}
        \caption{\emph{BaseKitch}}
    \end{subfigure}
    \begin{subfigure}[t]{0.32\textwidth}
        \includegraphics[width=\textwidth]{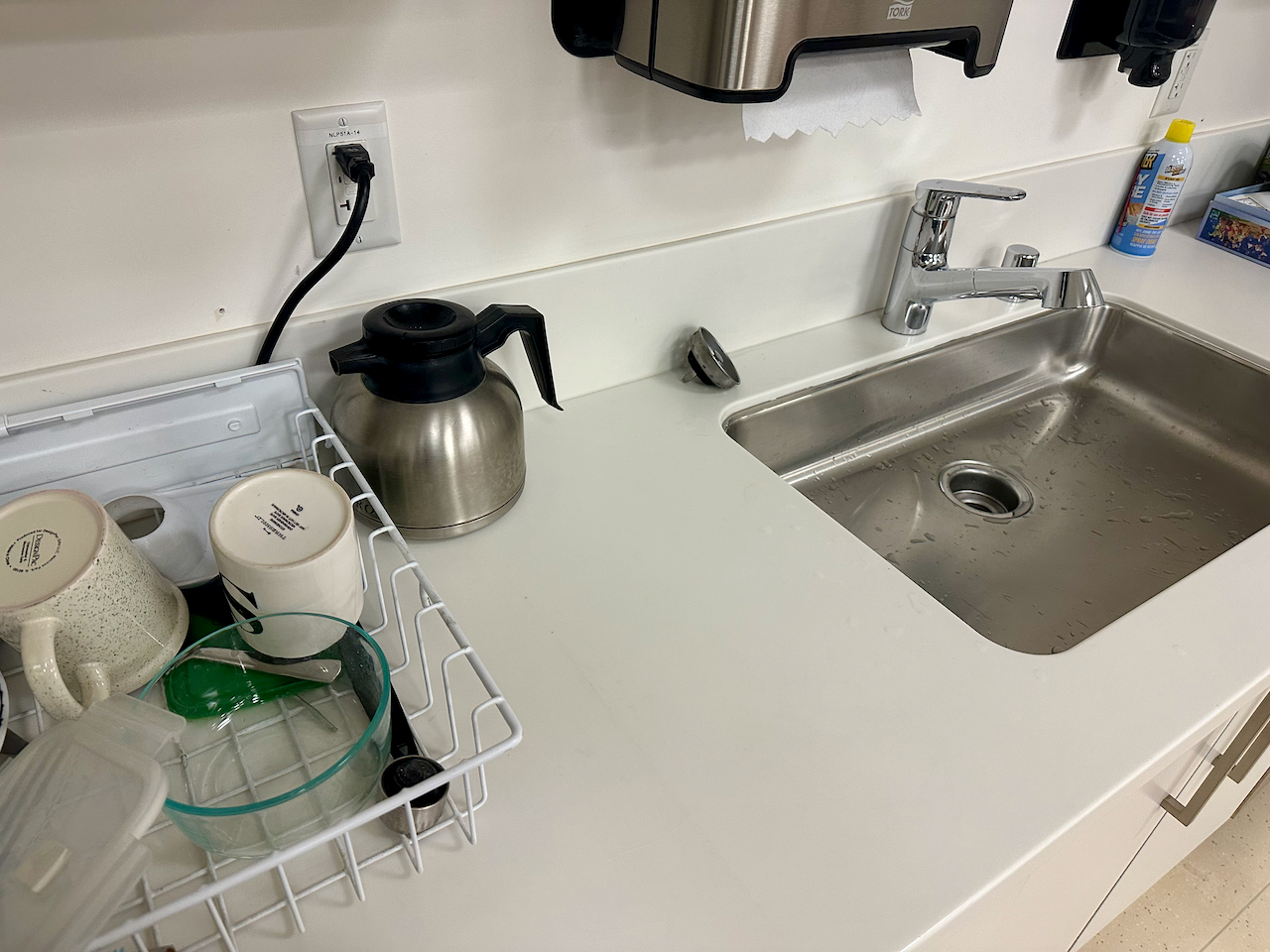}
        \caption{\emph{CompKitch}}
    \end{subfigure}
    \begin{subfigure}[t]{0.32\textwidth}
        \includegraphics[width=\textwidth]{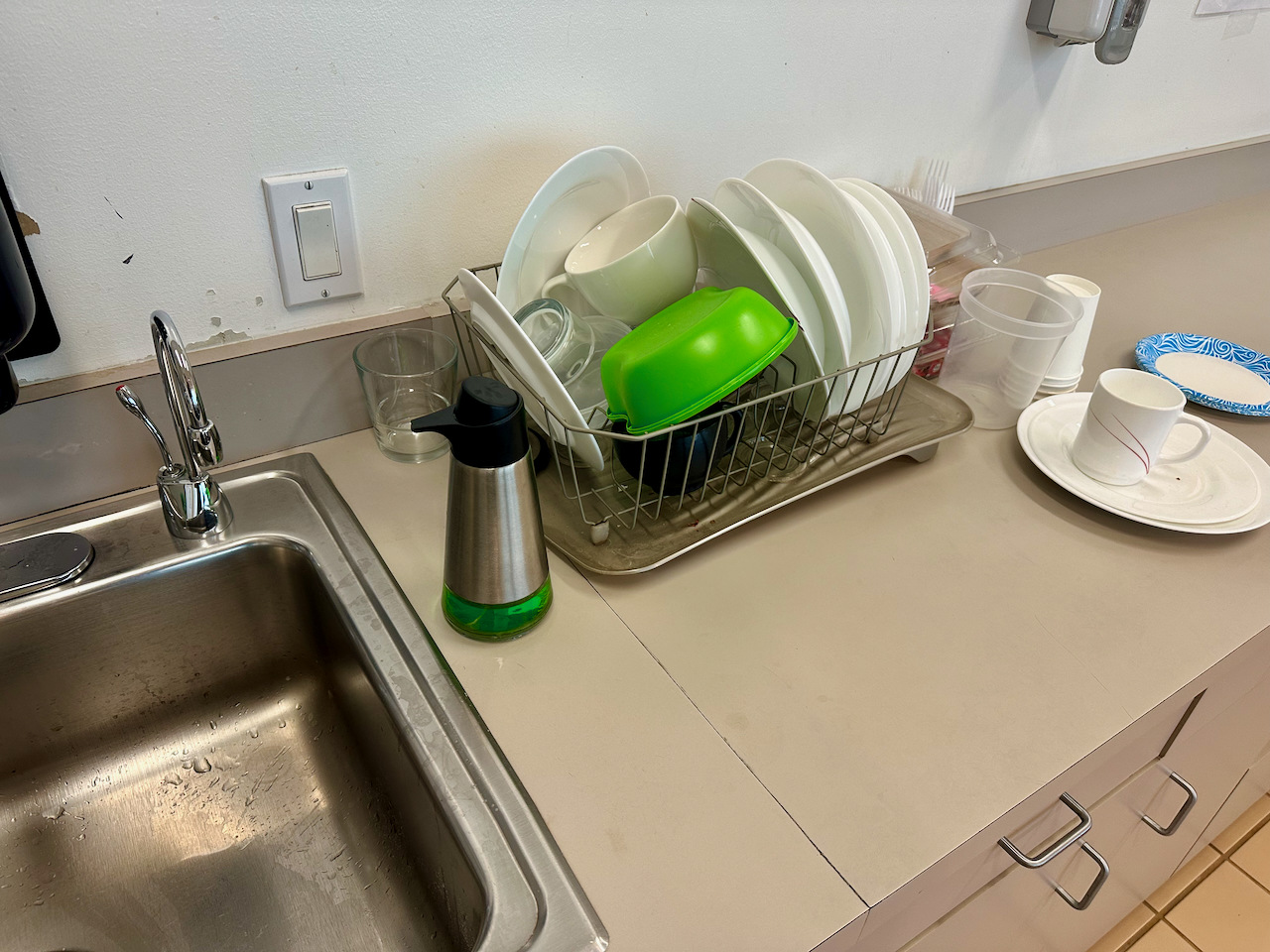}
        \caption{\emph{TileKitch}}
    \end{subfigure}
    \caption{\small{Additional views of each kitchen in our real robot experiments. We name \emph{CompKitch} and \emph{TileKitch} after their countertop materials of composite and tile, respectively. These images were taken after our evaluations, so there may be some slight differences from then.}}
    \label{fig:kitchen_views}
    \vspace{-17.5pt}
\end{figure*}

\subsection{Robot Platform}
\label[appendix]{sec:robot_platform_details}
We use Logitech C920 webcams for our main and secondary third-person cameras. Our setup also includes the wrist camera used in BridgeData V2, and we experimented with using it in addition to our main third-person camera. While we found that it improved overall robustness when training from scratch, particularly for some factors that wrist cameras provide invariance to (e.g., \emph{object position}, \emph{table height}), policies still benefited from data coverage for these factors. More importantly, we also found that it was incompatible with using BridgeData V2 as prior data, reducing performance when using prior data compared to not using prior data at all. We believe this is because only a small fraction of BridgeData V2 contains wrist camera data, as similar negative results have been observed in other work that use prior robotic datasets with a small proportion of wrist camera data \citep{octo_2023}. As we were only able to achieve successful transfer in our experiments in \cref{sec:ood_transfer} by using prior data without a wrist camera, we decided to omit the wrist camera in all our experiments.

\vspace{-4pt}
\subsection{Evaluation Protocol}
\label[appendix]{sec:real_eval_details}
For our out-of-domain transfer experiments in \cref{sec:ood_transfer}, we name our transfer kitchens \emph{CompKitch} and \emph{TileKitch} after their countertop materials of composite and tile, respectively. We provide additional views of these kitchens, as well as \emph{BaseKitch}, in \cref{fig:kitchen_views} to make the difference between these kitchens more apparent. These images were taken after our evaluations, so there may be some slight differences from then.

Our factor \emph{table height} refers to the height of the object table (where objects are manipulated on) relative to the robot. However, we vary this factor in practice by adjusting the height of the mobile table the robot and third-person cameras are mounted on. This still changes the relative height of the robot with respect to the object table, the same as if the object table changed height. For example, in \cref{sec:ood_transfer}, \textbf{Higher Table} refers to the object table being higher relative to the robot, which was achieved by lowering the robot's table.

We collect demonstrations using a Meta Quest 2 VR headset for teleoperation. All demonstrations are collected by a single experienced human teleoperator for consistency. We collect our \textbf{Stair} dataset by starting at $f^*$, and then varying factors cyclically in the order \emph{object position}, \emph{object type}, \emph{container type}, \emph{table height}, \emph{table texture}. The order we vary values for each factor is the same as in \cref{fig:real_factors}, from top to bottom.

\vspace{-3pt}
\subsection{Training}
\label[appendix]{sec:real_train_details}
We use the same ResNet-34 diffusion goal-conditioned policy architecture from the original BridgeData V2 experiments, except we condition on a history of 2 128x128 RGB image observations from a third-person view (in addition to a goal image from the same view), and use action chunking to predict the next 4 actions. However, we noticed better performance during inference by only executing the first predicted action, so we do this for all experiments unless otherwise stated. We share the same visual encoder across image observations in the history. We use the same 7D action space as the original experiments (6D end-effector pose deltas, and open/close gripper). We use the same data augmentation from the original experiments, which consists of random crops, random resizing, and color jitter.

For policies using BridgeData V2 as prior data, we first pre-train a model on BridgeData V2 using the original training hyperparameters for 2M gradient steps. As done in the original BridgeData V2 experiments, 10\% of trajectories in this data is reserved for validation. We checkpoint every 50K steps, and choose the checkpoint with the lowest validation action prediction mean-squared error as the initialization for later fine-tuning. During fine-tuning, we use the same hyperparameters as during pre-training, except we train for only 300K gradient steps. Also, we do not use a validation set, and instead simply evaluate the final checkpoint. When co-fine-tuning, we train on a mixture of 75\% in-domain data and 25\% prior data.

\begin{table*}[t]
    \begin{adjustbox}{width=\textwidth}
    \begin{tabular}{l|ccccc|ccccc}
        \toprule
        Data Strategy        & \multicolumn{10}{c}{\textbf{L}}  \\
        \midrule
        Train Method         & \multicolumn{5}{c|}{Bridge}                                                    & \multicolumn{5}{c}{From Scratch}                \\
        \midrule
        \diagbox{Factor 1}{Factor 2}
                             & \makecell{Object\\Pos} & \makecell{Object\\Type} & \makecell{Container\\Type} & \makecell{Table\\Height} & \multicolumn{1}{l|}{\makecell{Table\\Tex}}
                             & \makecell{Object\\Pos} & \makecell{Object\\Type} & \makecell{Container\\Type} & \makecell{Table\\Height} & \makecell{Table\\Tex} \\
        \midrule
        Object/Container Pos & N/A          & \textbf{7/9} & \textbf{6/9} & \textbf{2/9} & \multicolumn{1}{c|}{\textbf{6/9}} & N/A & 5/9  & 2/9 & 0/9 & 3/9     \\
        Object Orientation   & \textbf{2/9} & \textbf{5/9} & \textbf{5/9} & \textbf{1/9} & \multicolumn{1}{c|}{\textbf{3/9}} & 0/9 & 2/9  & 3/9 & 0/9 & 1/9     \\
        \midrule
        Overall              &  \multicolumn{5}{c|}{\textbf{36/81}}                                            & \multicolumn{5}{c}{16/81}                      \\
        \bottomrule
    \end{tabular}
    \end{adjustbox}
    \caption{\small{Additional real robot pairwise composition results for our ``\emph{put fork in container}" task, with additional factors \emph{object/container position} and \emph{object orientation}. Similarly as in our main pairwise composition results in \cref{tab:pairwise_results}, leveraging BridgeData V2 as prior data significantly improves composition for these factors compared to training from scratch.}}
    \label{tab:extra_results}
    \vspace{-17.5pt}
\end{table*}

\begin{figure}[t]
    \vspace{5pt}
    \centering
    \includegraphics[width=\columnwidth]{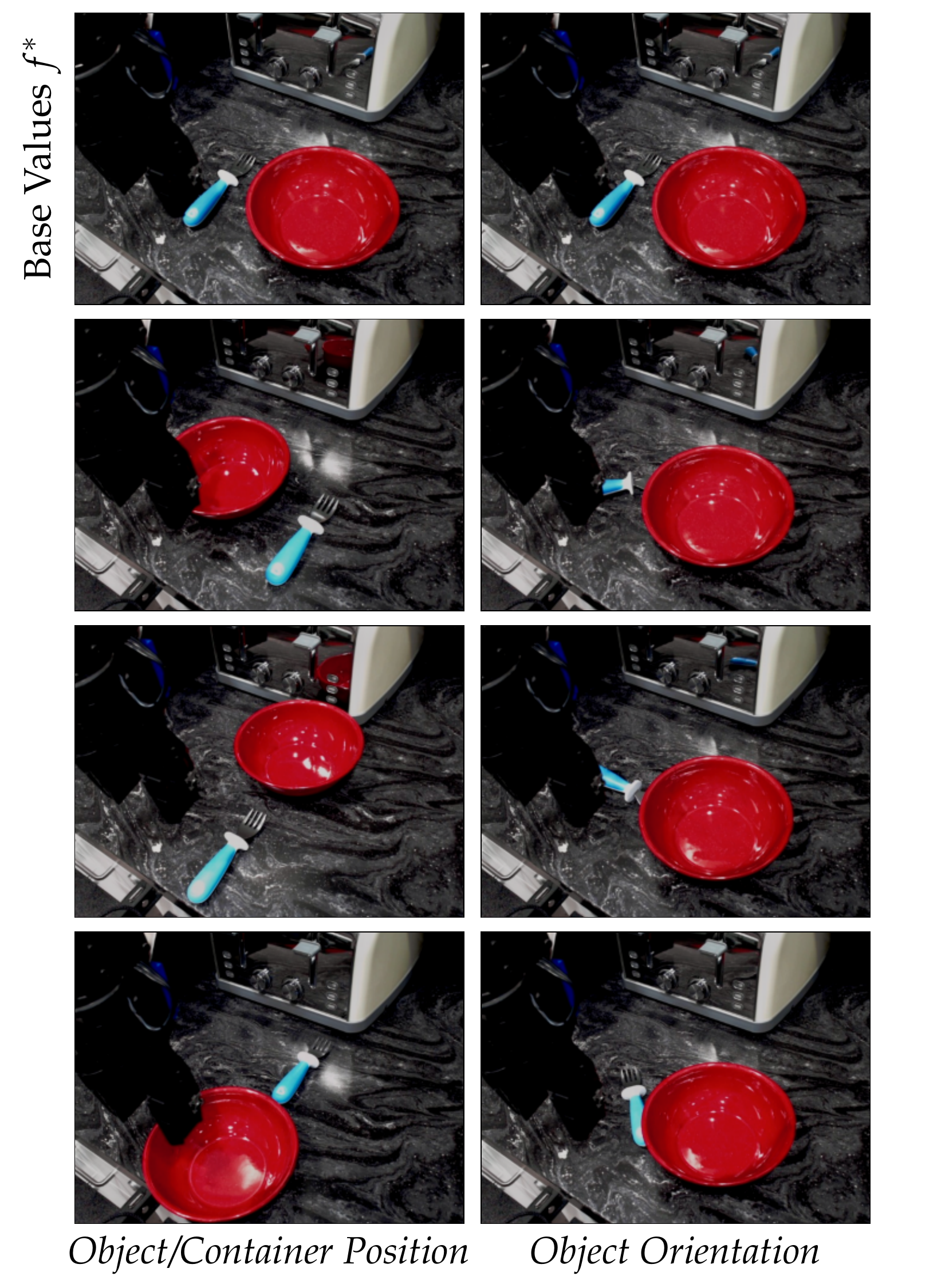}
    \vspace{-17.5pt}
    \caption{\small{Visualization of our additional real robot factors in \emph{BaseKitch}. The top row shows our base factor values $f^*$. The other rows show all deviations from $f^*$ by one factor value.}}
    \label{fig:extra_factors}
    \vspace{-20pt}
\end{figure}

\vspace{-2pt}
\subsection{Pairwise Composition for Additional Factors}
\label[appendix]{sec:additional_pairwise}

We conduct experiments for two new factors: \emph{object/container position} and \emph{object orientation}. We visualize these factors and their values in \cref{fig:extra_factors}.  Note that \emph{object/container position} involves new positions of both the fork and container, with more significant changes compared to our previous values for \emph{object position}. We extended the \textbf{L} dataset for our original factors with 10 demonstrations for each new factor value, re-trained policies on this extended dataset, and evaluated pairwise composition with these new factors. Unlike our previous results in \cref{sec:pairwise}, we do not evaluate the \textbf{No Variation} Bridge policy, because we found it was unable to succeed at all with shifts for these factors in isolation, so there was no potential for composition.

While these policies worked for \emph{object/container position}, we found they was unable to perform the task for the \emph{object orientation} shifts we considered in isolation. We hypothesize this could be because different values for this factor have relatively small changes in their visual observations, but require significantly different behavior, which can make it challenging to learn when to apply the correct behavior. Therefore, for our evaluation on composing \emph{object orientation}, we trained separate policies where we balance training batches such that 50\% of in-domain data consists of data for \emph{object orientation}.

We report these results in \cref{tab:extra_results}. We find that with prior data, \emph{object/container position} achieves similar composition as our original \emph{object position} factor, with a success rate of \textbf{20/36} for both. However, when training from scratch, the policy is able to compose this new factor more effectively than the original, achieving a success rate of \textbf{10/36} compared to \textbf{1/36}. This could be because different values for \emph{object/container position} are more visually distinguishable, and their required behavior is also significantly more different, which could make it easier to learn when to apply the correct behavior.

Our policies can sometimes compose \emph{object orientation}, although composition for this is the weakest compared to the other factors we study. This could be due to the aforementioned challenges with learning for this factor, as well as because our data balancing may have insufficiently represented the other factors.

\begin{figure*}[t]
    \centering
    \includegraphics[width=\textwidth]{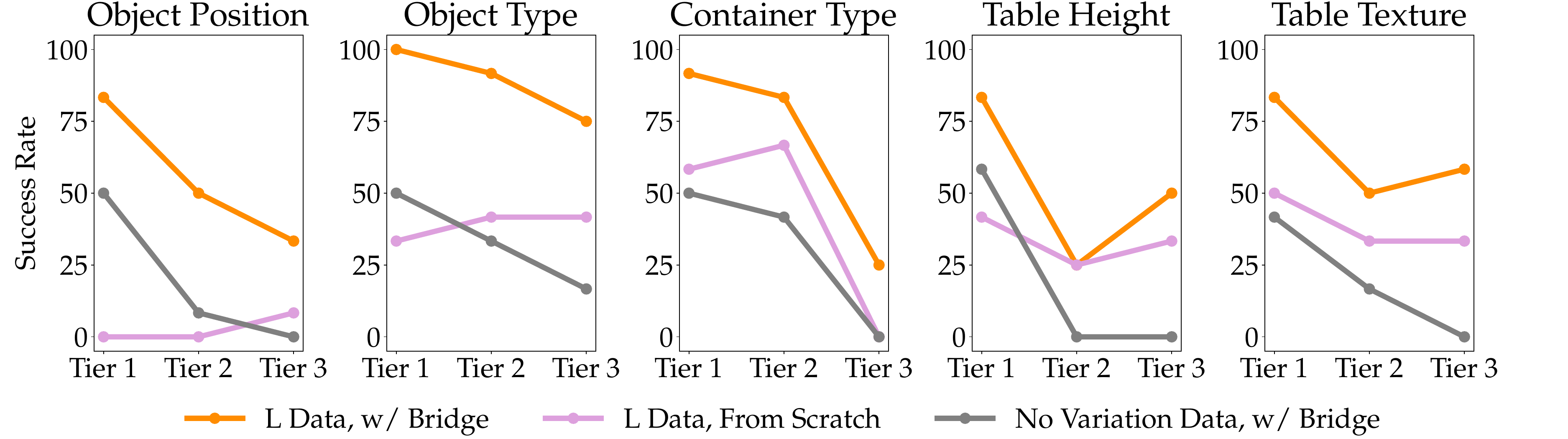}
    \vspace{-15pt}
    \caption{\small{Per-factor value success rates of policies from our pairwise composition results in \cref{tab:pairwise_results}}. Factor values are placed in tiers, where increasing tiers are more dissimilar from the base factor values $f^*$. Composition is generally more challenging for factor values that are more dissimilar from the base factor values.}
    \label{fig:similarity_analysis}
    \vspace{-17.5pt}
\end{figure*}

\vspace{-4pt}
\subsection{Factor Similarity Analysis}
\label[appendix]{sec:similarity_analysis}
Here, we provide additional analysis on when our policies are able to compose factor values from our pairwise evaluation in \cref{sec:pairwise}. To do this, for each factor, we consider how similar each of the non-base factor values we consider are to the base factor value, with respect to a policy's ability to generalize across factor values.

To compute a similarity metric, we consider the success rates from our results in \cref{tab:pairwise_results}, in particular for the policy trained on \textbf{No Variation} data and BridgeData V2. We then obtain a success rate for each non-base factor value, by aggregating the results for all factor value pairs that contain that factor value. We use this success rate as our similarity metric, where higher success rates indicate greater similarity, because this captures how well a policy trained on base factor values generalizes to other factor values.

\begin{table}[t]
    \centering
    \vspace{5pt}
    \begin{tabular}{l|lll}
        \toprule
        Factor         & Tier 1         & Tier 2        & Tier 3           \\
        \midrule
        Object Pos     & Down (3)       & Up (4)        & Left (2)         \\
        Object Type    & Wooden (3)     & Gray (4)      & Plastic (2)      \\
        Container Type & Blue Plate (4) & Pink Bowl (3) & White Cup (2)    \\
        Table Height   & Higher 5cm (2) & Lower 5cm (3) & Lower 8cm (4)    \\
        Table Tex      & Brown Wood (2) & Gift Wrap (4) & White Marble (3) \\
        \bottomrule
    \end{tabular}
\caption{\small{Tiers for each factor value used in \cref{fig:similarity_analysis}, determined using our success rate-based similarity metric as described in \cref{sec:similarity_analysis}}. In parentheses next to each factor, we provide the row number where the factor value is visualized in \cref{fig:real_factors}.}
\label{tab:factor_tiers}
\vspace{-20pt}
\end{table}

Using this similarity metric, we rank the non-base factor values for each factor as \emph{Tier 1}, \emph{Tier 2}, or \emph{Tier 3}, where a higher tier is more dissimilar from the base factor value. We list the tiers for each factor value in \cref{tab:factor_tiers}. We then take the aggregated per-factor value success rates for each policy from our results in \cref{tab:pairwise_results} (where we aggregate success rates as we do for computing the similarity metric), and plot this against our factor value tiers in \cref{fig:similarity_analysis}.

We find that the policies trained on \textbf{L} data (orange and pink lines) generally achieve lower compositional success rates for factor values that are more dissimilar from the base factor values $f^*$, suggesting that composition is more challenging for these more dissimilar values, although this trend is not strictly monotonic. We note that part of this effect could be because our success rate-based similarity metric may also capture how challenging factor values are in general.

\subsection{R3M}
We use the ResNet-50 version of R3M. When training, we pre-compute representations beforehand, and standardize the dataset such that each feature has mean 0 and standard deviation 1 (similar to the batch normalization used in the original R3M experiments). We use the training dataset mean and standard deviation for normalization during inference. We feed normalized representations to the same diffusion policy head architecture used when learning end-to-end, except we do not use goal-conditioning. We increase the amount of training gradient steps from 300K to 500K, to reduce training loss. We also tried 1M gradient steps, which reduces loss even further, but this resulted in worse performance. Instead of executing only the first predicted action as with the end-to-end policies, we execute all 4, which we found to slightly reduce jitteriness. Like when learning end-to-end, we verify our R3M policies are able to succeed with base factor values $f^*$ in \emph{BaseKitch}.

\subsection{VC-1}
We train and evaluate VC-1 policies using the same procedure as with R3M, except using the ViT-L version of VC-1 instead of R3M.

\end{document}